\RequirePackage[hyphens]{url}
\documentclass[11pt]{article} 
\usepackage[margin=1.05in]{geometry}
\usepackage{setspace}
\usepackage{amsmath,amssymb,amsfonts}
\usepackage{natbib}
 \bibpunct[, ]{(}{)}{,}{a}{}{,}%

\usepackage{rotating}
\usepackage{fancyvrb}
\usepackage{multirow}
\usepackage{array}
\usepackage{algorithm}
\usepackage{algpseudocode}
\usepackage{setspace}
\usepackage[export]{adjustbox}
\usepackage{anyfontsize}
\usepackage{bm}
\usepackage[toc,page, title, titletoc]{appendix}
\usepackage{longtable}
\usepackage{caption} 
\begin{document}

\title{Dynamic Demand Management for Parcel Lockers\vspace{-4ex}}
\date{\vspace{-4ex}}
\maketitle
\begin{center}
Daniela Sailer$^1$, Robert Klein$^1$, Claudius Steinhardt$^2$ \par
\footnotesize
daniela.sailer@wiwi.uni-augsburg.de, robert.klein@wiwi.uni-augsburg.de, \par
claudius.steinhardt@unibw.de
\end{center}
\begin{center}
\footnotesize
  $^1$ Chair of Analytics \& Optimization, University of Augsburg \par  
  $^2$ Chair of Business Analytics \& Management Science, University of the Bundeswehr Munich
\end{center} 

\begin{abstract}
    In pursuit of a more sustainable and cost-efficient last mile, parcel lockers have gained a firm foothold in the parcel delivery landscape. To fully exploit their potential and simultaneously ensure customer satisfaction, successful management of the locker’s limited capacity is crucial. This is challenging as future delivery requests and pickup times are stochastic from the provider’s perspective. In response, we propose to dynamically control whether the locker is presented as an available delivery option to each incoming customer with the goal of maximizing the number of served requests weighted by their priority. Additionally, we take different compartment sizes into account, which entails a second type of decision as parcels scheduled for delivery must be allocated. We formalize the problem as an infinite-horizon sequential decision problem and find that exact methods are intractable due to the curses of dimensionality. In light of this, we develop a solution framework that orchestrates multiple algorithmic techniques rooted in Sequential Decision Analytics and Reinforcement Learning, namely cost function approximation and an offline trained parametric value function approximation together with a truncated online rollout. Our innovative approach to combine these techniques enables us to address the strong interrelations between the two decision types. As a general methodological contribution, we enhance the training of our value function approximation with a modified version of experience replay that enforces structure in the value function. Our computational study shows that our method outperforms a myopic benchmark by 13.7\% and an industry-inspired policy by 12.6\%.
\end{abstract}


\section{Introduction} \label{intro}

Parcel lockers have evolved into a viable alternative to traditional home delivery and are gaining increasing popularity worldwide. Recent surveys show, e.g., that 54\% of online shoppers in Poland prefer to receive their parcels via parcel lockers \citep{pref_poland}, 57\% in China \citep{pref_china}, and in Estonia, 70\% favor parcel lockers over other delivery options such as home delivery \citep{pref_estonia}. As these lockers offer a huge potential to consolidate deliveries and decrease failed delivery attempts \citep{savelsbergh_van-woensel2016}, this is an encouraging development in the quest for a green and cost-efficient last mile. 

In practice, the introduction of parcel lockers has given rise to three general business models that vary in the degree of integration across different steps of the delivery process \citep[cf.][]{oohd_survey}: Firstly, \textit{locker providers} like myflexbox, Hive Box, or Pick Network focus entirely on operating the locker infrastructure and renting out locker capacity to other companies. Secondly, in addition to maintaining proprietary locker infrastructure, \textit{logistics service providers}, such as DHL, Australia Post, or InPost, perform the delivery to the lockers. Lastly, taking this even a step further, \textit{retailers} like Amazon or Alibaba with its logistics-subsidiary Cainiao do not only own parcel locker networks and perform delivery services, but also act as parcel senders to ship the goods they sell to the end consumer.

In all three cases, parcel recipients play a pivotal role in two ways: On the one hand, their purchasing habits and preferences for different parcel locker locations shape the number of incoming delivery requests for a specific locker. Generally, recipients tend to prefer lockers that are close to their home or that can be easily integrated into regular trips, e.g., on the commute to work. In practice, recipients usually select a specific locker as the desired delivery address. On the other hand, the speed at which recipients pick up their parcel directly affects the available capacity over time. As a result, recipients' behavior creates two sources of uncertainty from a locker operator's perspective.

Due to these stochastic influences, parcel lockers are inherently susceptible to short-term mismatches between capacity and the number of incoming requests. Typically, a parcel locker consists of individual compartments of different sizes. This introduces compatibility constraints as the compartment assigned to a specific parcel must be large enough. The total number of compartments per size and thus the locker's capacity is fixed in the short term. By contrast, the availability of compartments over time is stochastic as it depends on how quickly recipients pick up their parcels. For a successful delivery, a compatible compartment must be available from the time of a parcel's delivery up until the recipient collects it from the locker or the maximum storage time (usually between three to seven days) is exceeded. 

In reality, it may regularly occur that parcels cannot be delivered to the locker chosen by the recipient because all compatible compartments are already occupied. As a widespread strategy, companies like DHL redirect parcels to another locker in such cases. However, this may leave the recipient very dissatisfied, especially if picking up their parcel from the alternative locker requires a large detour. Furthermore, depending on the business model, recipients can be grouped according to different priority levels resulting from shipping options such as same-day, next-day and standard delivery \citep{sethuraman_etal} or negotiated contracts and service-level agreements with the locker/logistics service providers \citep{dhl_SLAs}. If the parcels of high-priority customers tend to arrive on short notice (e.g., because of express shipping), the problem is exacerbated as they might get redirected disproportionately often. 

Consequently, we advocate actively managing scarce locker capacity instead of merely handling capacity shortages in a reactive way. To accomplish this, we propose to \textit{dynamically control the availability} of the parcel locker as a delivery option for each incoming request. We thereby eliminate the need to redirect parcels and are simultaneously able to reserve capacity for high-priority recipients. We call the resulting problem the \textit{Dynamic Parcel Locker Demand Management Problem} (DPLDMP). It can be applied to any of the three business model settings described above. For ease of exposition, we refer to the decision maker as a general `provider' and use the terms `customer' and `recipient' synonymously.  

More specifically, the DPLDMP can be summarized as follows: We focus on a single parcel locker and assume that parcels are delivered to the locker once per day. The locker consists of multiple compartments that differ in their size. The availability of compartments over time depends on the customers' pickup behavior and is uncertain. Customers are divided into groups with given priority levels as a result of contracts, shipping options, or service level targets. Over the course of each day, \textit{requests} arrive stochastically. For each incoming request, the provider determines whether a suitable compartment is guaranteed to be available at the time of delivery to eliminate the need for fallback measures such as redirecting parcels to other lockers. Subsequently, the provider makes a \textit{demand control decision}, i.e., decides whether to accept or reject the request. In the case of acceptance, the request turns into an \textit{order} and the associated parcel has to be delivered to the locker on its given delivery date. If the provider rejects the request, the customer gets informed that the parcel locker is currently not available as a delivery option. The customer might then choose an alternative option such as home delivery, which is outside of our paper's scope. At the end of each day, after a fixed cut-off time, the provider assigns the parcels scheduled for delivery on the following day to the available locker compartments, which we refer to as the \textit{allocation} decision. The overall goal is to maximize the expected number of accepted customer requests weighted by their priority.

Due to the stochastic influences and dynamic decision making, the DPLDMP is an infinite-horizon sequential decision problem. Finding a good policy for it is difficult for two main reasons:
\begin{enumerate}
    \item \textit{Two types of strongly interrelated decisions.} In our problem, the provider has to make two types of nontrivial decisions, demand control and allocation, at different points in time. These decisions are heavily intertwined: Demand control determines which requests turn into orders and must then be allocated to locker compartments later on. In the same vein, allocation decisions affect which compartments are occupied, thereby shaping the decision space for future demand control.
    \item \textit{Displacement effects.} Accepting a request yields an immediate reward as the number of accepted requests increases. At the same time, the available capacity decreases such that the provider might not be able to accept a higher-priority request arriving in the future. Similarly, an inefficient allocation of parcels to compartments can impair subsequent demand control. Both decision types thus entail displacement effects that one has to anticipate and, in the case of demand control, trade off with the immediate reward. These displacement effects are hard to estimate because future incoming requests as well as the capacity consumption of requests and orders are uncertain.
\end{enumerate}

To address these challenges, we develop a solution framework that combines and orchestrates multiple techniques from Sequential Decision Analytics \citep{powell2022} and Reinforcement Learning \citep[RL,][]{sutton+barto}. At its core, we employ cost function approximation (CFA) for the allocation decision and an offline trained parametric value function approximation (VFA) with a truncated online rollout \citep{bertsekas2022} for demand control. Applying these methods to the DPLDMP is not straightforward and requires adaptions specific to our setting. Notably, our approach needs to cope with two types of strongly entangled decisions at different points in time. To achieve this, we design the CFA such that it prioritizes creating favorable conditions for demand control. For the VFA, we incorporate uncertain pickups and anticipated allocation decisions for existing orders in a tentative allocation plan, which forms the basis for our features. As a result, we take into account the impact of one decision type on the other and vice versa.

With our work, we contribute to the literature in the following ways:
\begin{itemize}
    \item To the best of our knowledge, we are the first to consider demand management for parcel lockers with multiple compartment sizes and formalize it as a sequential decision problem.
    \item We propose an anticipatory solution framework tailored to our problem that handles two types of decisions at different points in time in an infinite horizon. Importantly, we explicitly address the strong interrelations between the decision types by combining different algorithmic techniques such as CFA and VFA in a novel way that embeds the interdependencies of decisions in the design of the individual components.
    \item As a general methodological contribution, we augment an existing temporal difference learning method using the principle of experience replay to learn the weights for the VFA. This modification stabilizes the learning process and allows us to enforce structural properties in the value function. Our approach offers a novel way to include domain knowledge in VFA and presents a promising starting point for future work.
    \item Throughout our paper, we present a total of seven key insights into the DPLDMP gained from theoretical analysis and our computational study. Among them, we find that our solution approach beats both a myopic benchmark and an industry-inspired policy, we investigate under which conditions anticipation is particularly worthwhile, and we show that customers can be prioritized both explicitly and implicitly.
\end{itemize}

The remainder of this paper is structured as follows: In Section \ref{related_work}, we review the literature related to the DPLDMP and highlight the resulting research gap. In Section \ref{problem_statement}, we provide a detailed problem description along with a discussion of key assumptions and model the DPLDMP as a sequential decision problem. In Section \ref{solution_approach}, we propose our solution framework. We present the results of our computational study in Section \ref{comp_study} and conclude the paper with future research directions in Section \ref{conclusion}.

\section{Related Work} \label{related_work}

In this section, we give a concise review of the three major literature streams related to the DPLDMP. Section \ref{rw_OOHD} focuses on demand management for out-of-home delivery, thereby covering publications in a similar application context. Taking on a more theoretical perspective, Section \ref{rw_reusable_resources} is dedicated to relevant publications on demand management with reusable resources, and Section \ref{rw_upgrading} addresses research involving upgrades. We summarize our findings with a tabular overview and delineate the paper at hand from the most closely related existing work in Section~\ref{rw_summary}.

\subsection{Demand Management for Out-of-Home Delivery} \label{rw_OOHD}
The last leg of the parcel delivery process, also known as the last mile, is notoriously cost-intensive, incentivizing  researchers and practitioners to innovate. This manifests itself in two key trends: First, demand management has been integrated into various existing logistics services such as attended home delivery or same-day delivery in recent years \citep{fleckenstein_etal_2023}. By controlling the availability or prices of delivery options, providers can actively steer demand and increase profitability \citep{waßmuth_etal_2023}. Second, various novel delivery concepts have emerged \citep{boysen_etal_2021}. A prominent example is out-of-home delivery (OOHD), i.e., the delivery to parcel lockers or parcel shops instead of the recipient's home address. The DPLDMP is directly at the intersection of these two trends.

The review by \cite{oohd_survey} documents a surge in publications on OOHD, especially for location planning and routing problems. By contrast, research on demand management for OOHD is still in its nascent stage. While there is work on the satisfaction of customer preferences on an aggregate level \citep{dumez_etal_2021} and OOHD product design \citep{janinhoff_etal_2023}, our analysis focuses on publications that explicitly take into account operational demand management, thereby also excluding work on dynamic OOHD problems without any demand steering component \citep[e.g.,][]{ulmer+streng_2019}. To the best of our knowledge, there exist only three publications matching these criteria, which we present in the following.

\cite{galiullina_etal_2024} consider demand management at the start of a retailer's fulfillment planning phase where all orders due for delivery on a specific day are known. The default delivery location is the home address, but the retailer can offer selected recipients a monetary incentive to switch to OOHD. Intervening at an earlier stage, \cite{akkerman_etal_2023} investigate joint availability control and dynamic pricing of delivery options during the order arrival phase for a single delivery day. In cooperation with retailer Amazon, \cite{sethuraman_etal} study availability control for a parcel locker with uniform compartments.

\subsection{Reusable Resources} \label{rw_reusable_resources}

From a theoretical perspective, the DPLDMP can be cast as a revenue management problem with reusable resources. Basically, revenue management is a special case of demand management where costs can either be neglected or attributed to individual customers \citep{waßmuth_etal_2023}. In a setting with reusable resources, customers arrive dynamically and, in case of a purchase, make use of a resource for a certain (potentially stochastic) amount of time. Afterward, the resource returns to the seller and can thus be allocated several times over the course of the planning horizon. For parcel lockers, resources correspond to the compartments, and the usage duration is stochastic as it depends on how quickly the recipient picks up the parcel. Moreover, there are different sizes of compartments, i.e., multiple types of resources, and delivery requests typically arrive some time ahead before the actual delivery. We hence focus on publications that consider stochastic usage durations in combination with a) multiple resource types or b) advance reservations as this most closely matches our problem setting.

\textit{Multiple resource types.} \cite{püschel_etal_2015} investigate the effectiveness of dynamic pricing compared to availability control as well as a first-come-first-served policy for a cloud-computing provider. Jointly controlling availability and prices, \cite{owen_simchi-levi_2018} compare static and dynamic policies. Extending this, \cite{rusmevichientong_etal_2020} combine a static policy with a rollout to obtain a dynamic policy that takes the current resource utilization into account. \cite{gong_etal_2022} show that under certain assumptions for the choice probabilities and usage durations, a simple policy that always offers the myopically optimal assortment provably earns at least half the expected cumulative revenue of an optimal clairvoyant benchmark. \cite{baek_ma_2022} generalize availability control with reusable resources to a network setting where a product may require a combination of multiple resource types.

\textit{Advance reservations.} \cite{papier_thonemann_2010} study availability control for a cargo rail company and propose an anticipatory policy that exploits the information revealed by advance reservations.

\subsection{Upgrading} \label{rw_upgrading}
If alternatives can be ordered in a hierarchy, a seller can make use of upgrading by satisfying the demand for a specific product with another product higher in the hierarchical order. This enables the seller to alleviate short-term mismatches between capacity and demand. In the context of parcel lockers, this mechanism is relevant due to the compartment sizes, i.e., a parcel can be `upgraded' to a larger compartment size than strictly necessary. While customers generally do not care about the compartment size as long as it is compatible with their parcel, methodology-wise this still equals a setting with upgrades.  

Out of the corresponding literature stream, we focus on publications with full cascading upgrades, i.e., upgrades to any product higher in the hierarchical order. Given that the resulting papers exhibit largely identical characteristics with regard to our classification scheme in Section \ref{rw_summary}, we center the discussion around three representative publications and refer the interested reader to the review by \cite{gönsch_2020}. \cite{gallego_stefanescu2009} compare different upgrade mechanisms and incorporate fairness considerations. \cite{gönsch_steinhardt2015} apply availability control with upgrades to an airline network and show certain monotonicity properties of opportunity cost. In a very recent publication, \cite{zhu_topaloglu2024} study availability control for flexible products, which are a generalization of upgrades.
 
\subsection{Summary and Research Gap} \label{rw_summary}

\begin{table}
\caption{Related literature.\label{overview_related_work}}
\centering
\resizebox{\textwidth}{!}{\renewcommand{\arraystretch}{1.2}
\begin{tabular}{m{2cm} >{\raggedright}m{5.85cm} >{\centering\arraybackslash}p{1.5cm} >{\centering\arraybackslash}p{1.7cm} >{\centering\arraybackslash}p{1.7cm} >{\centering\arraybackslash}p{2cm} >{\centering\arraybackslash}p{1.8cm} >{\centering\arraybackslash}p{1.5cm} >{\centering\arraybackslash}p{1.5cm}} \hline
 & \multirow{2}{*}{Publication} & \multirow{2}{*}{Decision} & Multiple resources & \multirow{2}{*}{Upgrades} & Advance reservations & Stochastic duration & Infinite horizon & Solution concept \\ \hline
 \multirow[t]{3}{*}{OOHD} & \cite{akkerman_etal_2023} & AC, P & \checkmark & & \checkmark & & & A \\
& \cite{galiullina_etal_2024} & P & \checkmark & & & & & A \\
& \cite{sethuraman_etal} & AC & & & \checkmark & \checkmark & \checkmark & DLP \\
\hline
Reusable & \cite{baek_ma_2022} & AC & \checkmark & & & \checkmark &  & CDLP \\ 
\multirow[t]{6}{*}{Resources} & \cite{gong_etal_2022} & AC & \checkmark & & & \checkmark &  & M \\
& \cite{owen_simchi-levi_2018} & AC, P & \checkmark & & & \checkmark & \checkmark & CDLP \\
& \cite{papier_thonemann_2010} & AC & & & \checkmark & \checkmark & \checkmark & A \\
& \cite{püschel_etal_2015} & AC, P & \checkmark & & & \checkmark &  & A \\
& \cite{rusmevichientong_etal_2020} & AC, P & \checkmark & & & \checkmark &  & A \\
\hline
\multirow[t]{3}{*}{Upgrading} & \cite{gallego_stefanescu2009} & AC, P & \checkmark & \checkmark & \checkmark & & & DLP \\ 
& \cite{gönsch_steinhardt2015} & AC & \checkmark & \checkmark & \checkmark & & & DLP \\ 
& \cite{zhu_topaloglu2024} & AC & \checkmark & \checkmark & \checkmark & & & A \\ 
\hline
& Our work & AC & \checkmark & \checkmark & \checkmark & \checkmark & \checkmark  & A \\
\hline
\multicolumn{9}{p{1.3\textwidth}}{\textit{Abbreviations.} Anticipatory (A), availability control (AC), choice-based deterministic linear program (CDLP), deterministic linear program (DLP), myopic (M), pricing (P).}
\end{tabular}}
\end{table}

Table \ref{overview_related_work} summarizes the literature related to the DPLDMP. We characterize the publications according to the following dimensions: The column `Decision' indicates the type of demand management decision (availability control (A) or pricing (P)). The next five columns track whether the problem features multiple types of resources, upgrading, advance reservations, stochastic usage durations, and an infinite planning horizon. The last column classifies the solution approach as myopic (M) or anticipatory (A). As a special case of the latter, we highlight papers drawing on the deterministic linear program (DLP) or its choice-based variant (CDLP) because this is a fundamental revenue management concept \citep{gallego+topaloglu}.

In line with the reviews by \cite{ma_etal_2022} and \cite{oohd_survey}, we find that demand management in OOHD is an emerging topic addressed by only a handful of very recent publications. For all of them, one major line of distinction to our work is how capacity is modeled. \cite{akkerman_etal_2023} and \cite{galiullina_etal_2024} assume uncapacitated facilities, while \cite{sethuraman_etal} consider a capacitated locker with uniform compartments. In contrast, we take into account multiple compartment sizes, thereby modeling capacity on a more granular and realistic level.

Although problems similar to the DPLDMP have been covered in the context of reusable resources or upgrading, our analysis reveals a substantial research gap. Essentially, there are no publications on problems with stochastic usage durations that simultaneously feature upgrading. Note that upgrading introduces an additional decision to the problem as customers need to be allocated to a specific resource type. Vice versa, papers which do take into account upgrading lack the stochastic usage duration that is central to OOHD as the available capacity depends on the recipients' pickup behavior. Our work fills this gap and may also be relevant to other application domains, e.g., to manage the demand for parking lots consisting of spaces with and without charging infrastructure for electric vehicles or station-based car sharing with multiple vehicle categories.

Concluding this section, we delineate our work from the two most closely related publications. On the one hand, the assumptions on usage durations prevent us from extending the methodology by \cite{papier_thonemann_2010} to the DPLDMP, especially as they do not allow for the distributions of durations to depend on the customer type. This contradicts the empirical observation that customers with faster shipping options tend to pick up their parcels more quickly \citep{sethuraman_etal}. On the other hand, the DPLDMP shares strong similarities with \cite{sethuraman_etal}, who use a DLP-based approach. The DLP is an established method and, as Table \ref{overview_related_work} shows, is also employed by a number of other publications. While we cannot directly apply the approach by \cite{sethuraman_etal} to our problem, we test a DLP-based benchmark adapted to our setting with multiple, upgradeable resources in our computational study.

\section{Problem Statement} \label{problem_statement}
In this section, we give a detailed description of the DPLDMP and formalize it as a sequential decision problem. Appendix \ref{overview_notation} provides a tabular overview of the notation. We conclude the section with an example.

\subsection{Problem Definition} \label{problem_description}
In short, the DPLDMP considers a single \textit{parcel locker} to which parcels are delivered once every day. The availability of the \textit{compartments} is uncertain and depends on the customer's \textit{pickup time}, which is in turn bounded by the \textit{maximum storage time}. New delivery \textit{requests} arrive stochastically over the course of each day. After performing a \textit{feasibility check}, the provider makes a \textit{demand control} decision, i.e., accepts or rejects the request. In case of acceptance, the request turns into an \textit{order}. At the end of each day, orders due for delivery are \textit{allocated} to locker compartments. The objective is to maximize the expected number of accepted requests weighted by their \textit{priority}. 

Section \ref{detailed_problem_description} introduces the main problem components on a more granular level together with the corresponding notation. In Section \ref{assumptions}, we discuss our key assumptions.

\subsubsection{Description and Notation.} \label{detailed_problem_description}

We define the setting of the DPLDMP as follows:

\textit{Resources.} We consider a single parcel locker. The locker consists of a limited number of compartments that differ in their size $\delta\in\mathcal{D}=\{1,\dots,D\}$. The compartment sizes are indexed in ascending order and allow full cascading upgrades, i.e., if a parcel requires a compartment of size $d\in\mathcal{D}$, it can be allocated to any available compartment of size $\delta\geq d$. The total number of compartments per size $\delta$ is denoted by $Q_\delta$.

\textit{Planning horizon.} The planning horizon is divided into days $\tau=1,\dots,\infty$. Each day is further discretized into sufficiently small time periods such that at most one request arrives per period. This yields a set of discrete points in time $t\in\mathcal{T}=\{1,\dots,T\}$ per day. Parcels are delivered to the locker once every day. For the allocation of parcels to compartments at the end of each day, we introduce an additional point in time $T+1$ with no request arrival. An entire day $\tau$ is thus represented by $\mathcal{T}^+=\mathcal{T}\cup\{T+1\}$.

\textit{Requests.} On each day, requests arrive dynamically over the course of $\mathcal{T}$ according to a known stochastic process $\mathfrak{F}^a$. Requests are characterized by three attributes: Firstly, each request belongs to a certain \textit{customer type} $c\in\mathcal{C}=\{1,\dots,C\}$. The customer type in turn corresponds to a \textit{priority weight} $m_{c}\in\mathcal{M}$ and, as described in the next paragraph, may correlate with how quickly customers pick up their parcels. Secondly, the \textit{parcel size} $d\in\mathcal{D}$ indicates that the request requires a compartment of size $\delta\geq d$. Thirdly, the \textit{lead time} $e\in\mathcal{E}=\{1,\dots,E\}$ determines when the parcel must be delivered to the locker. More precisely, a parcel with lead time $e$ has to be assigned to a compartment in exactly the $e$\textsuperscript{th} allocation decision from now if accepted. The lead time of each request is given, i.e., the delivery date cannot be pre- or postponed.

\textit{Parcel pickups.} The \textit{pickup time} $\psi$ refers to the amount of time after allocating a parcel to a compartment up until the compartment becomes available again. We express the pickup time as a tuple $\psi=(b,q)$ where $b$ represents the number of days after allocation and $q\in\mathcal{T}$ the point in time at which the compartment's status changes to unoccupied. Given that this depends on whether and when the customer collects the parcel, the pickup time is uncertain from the provider's perspective. It follows a known distribution $\mathfrak{F}^p$ that may depend on the customer type $c$. If the customer does not collect the parcel before a specific number of days $B$ - called the \textit{maximum storage time} - elapses, the provider removes the parcel from the locker. Consequently, the maximum storage time automatically yields an upper bound on the pickup time with $b\in\mathcal{B}=\{1,\dots,B\}$.

Having defined the basic setting, we elaborate on the provider's decision making:

\textit{Feasibility check.} For each request, the provider first determines whether the request is feasible. To classify a request as feasible, the decision maker must ensure that a sufficiently large compartment is going to be available from the time of delivery up until the customer picks up the parcel or the maximum storage time $B$ elapses. The availability of compartments hinges on the pickup times of orders currently in the locker as well as the orders scheduled for delivery in the next $E$ days as a result of previously accepted requests. Given that pickup times are uncertain, we hedge against all possible realizations of pickup behavior to guarantee a successful delivery. More precisely, we require a feasible allocation to exist even if all customers were to make use of the maximum storage time. We discuss the implications of this in the subsequent section.

\textit{Demand control.} After the feasibility check, the provider makes a demand control decision, i.e., accepts or rejects the request. In case of acceptance, the request turns into an \textit{order}. If the request is rejected, the customer may choose another delivery option such as home delivery, which is out of this paper's scope. 

\textit{Allocation.} At the end of each day ($t=T+1$), the provider allocates parcels scheduled for delivery on the following day to available and compatible compartments. Once a parcel is assigned to a specific compartment, it remains there until the customer collects it or the maximum storage time is reached. 

\textit{Objective.} The provider seeks to maximize the expected number of accepted requests weighted by $m_c$.

\begin{figure} 
\centerline{\includegraphics[scale = 0.9]{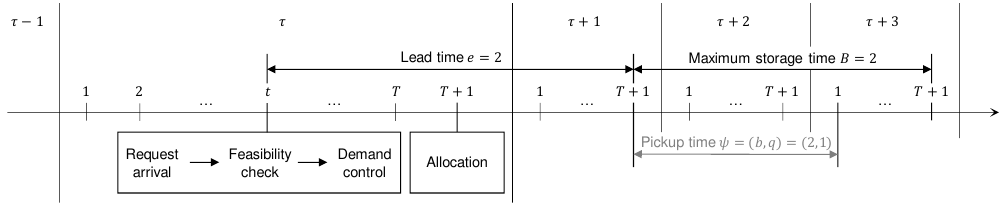}} 
\caption{Timeline for decisions and events. \label{figure_timeline}} 
\caption*{\footnotesize\textit{Note.} In the example, a request arrives at $t$ on day $\tau$ with lead time $e=2$. If the request is feasible and accepted, it has to be allocated in $T+1$ on day $\tau+1$ due to its given lead time. The customer may then collect the parcel starting with $t=1$ on $\tau+2$. The maximum storage time is $B=2$ and spans from the time of allocation at $T+1$ on $\tau+1$ up until $T+1$ on $\tau+3$, meaning that the compartment assigned to the parcel is guaranteed to be available again by then. Customers usually collect their parcel before the maximum storage time elapses, leading to a specific pickup time $\psi$. The pickup time is uncertain at the time of demand control. In the example, the request is accepted and the pickup occurs on day $\tau+3$ at $t=1$, leading to a pickup time of $\psi=(b,q)=(2,1)$.} 
\end{figure}

We illustrate the sequence of decisions and events along an exemplary timeline in Figure \ref{figure_timeline}.

\subsubsection{Main Assumptions.} \label{assumptions}

Our definition of the DPLDMP draws on the following assumptions:
\begin{enumerate}
    \item The assumption of locker compartments allowing full cascading upgrades aligns with the locker layout configurations commonly encountered in practice. Typically, parcel lockers consist of multiple tower modules of uniform width such that the compartments sizes merely differ in their height.
    \item For ease of exposition, one can imagine the actual delivery to the locker happening immediately after the allocation decision (for the parcels that are due for delivery on the subsequent day). Nevertheless, our problem definition does not rely on this simplification as long as the delivery is executed once each day and roughly at the same time. Consequently, we do exclude detailed operational planning such as vehicle routing and same-day delivery because it creates a highly complex setting in its own right, even without any demand steering component \citep[e.g.,][]{ulmer+streng_2019}. 
    \item The DPLDMP focuses on parcel lockers in a last-mile logistics context. In practice, parcel lockers additionally play an increasingly important role in first-mile logistics since parcel senders can also drop off their parcels (e.g., returns) at the locker to be then retrieved and further processed by the logistics service provider. We assume that the vehicles performing the delivery to the locker have sufficient capacity to handle all dropped-off parcels upon their daily visit, allowing us to neglect first-mile logistics.
    \item To represent the choice behavior of customers, we employ the independent demand model \citep{strauss_etal_2018}. In other words, we assume that every customer is interested only in one specific locker. This translates to the following process from the customer's perspective: Each customer considering delivery to a locker already has a preferred locker in mind when shopping online. During the checkout process, the customer selects the desired locker through an interface, which triggers a request arrival from the provider's perspective. The demand control decision can also be interpreted as whether the locker is displayed as available to the customer. In case of acceptance, the locker is displayed as available, and according to the independent demand model, the customer selects it as the desired delivery address for certain. In case of rejection, the customer chooses another delivery option such as home delivery or abandons the purchase. The assumption of independent demand is common in the OOHD literature, particularly as the special case of customers favoring the locker closest to their home \citep[e.g.,][]{galiullina_etal_2024}. While this is likely justified in sparse locker networks, it does neglect dynamic substitution effects that arise in denser networks where customers can easily switch to an alternative locker if their preferred one is unavailable.  
    \item The prioritization of customer types is an input parameter to be predetermined by the decision maker. It may result from the selected shipping option (e.g., standard versus express shipping), customer relationship management considerations, or service-level agreements \citep{dhl_SLAs}. Note that it also allows us to incorporate aspects specific to the business models introduced in Section \ref{intro}. In case of a logistics service provider, the priority weight could include an estimate of the cost to deliver the parcel to the customer's home address or a risk measure for failed deliveries. From a retailer's perspective, the priority weight might reflect the order basket value if the unavailability of the parcel locker risks the customer entirely abandoning the purchase. Consequently, setting the priority weights is an application-specific process and is governed by the overarching business strategy. We evaluate the impact of different prioritizations in Section \ref{comp_study}.
    \item We assume the size of the parcel to be known upon request arrival. As an example from practice, Amazon automatically tracks during the customer's checkout process whether an order is eligible for delivery to a locker due to its size, weight, and other factors \citep{amazon}, rendering this assumption realistic.
    \item The delivery date and hence the lead time are fixed and the provider cannot pre- or postpone deliveries. While more flexibility is conceivable in the case of a retailer, logistics service providers would first have to establish sufficient intermediate storage capacities. To keep the problem setting relevant for as many applications as possible, we treat the lead time as given.
    \item Empirical data shows that the customers with expedited shipping options tend to retrieve their parcels more quickly than those with standard shipping \citep{sethuraman_etal}. This motivates customer-type-dependent distributions of pickup times. The maximum storage time is a standard concept in practice, typically ranging between three to seven days, and applies uniformly to all customer types.
    \item With the feasibility check, we aim to eliminate the error of misclassifying an infeasible request as feasible. Orders wrongly classified as feasible create the need for fallback measures such as rerouting parcels to alternative lockers or undertaking multiple delivery attempts, which can cause substantial customer dissatisfaction. To avoid this, we must hedge against all possible realizations of pickup behavior. Consequently, to classify a request as feasible, we require a feasible allocation to exist even in the worst case where every customer makes use of the maximum storage time. Thereby, we can guarantee that the delivery to the locker is going to be successful, which resembles Amazon's communicated strategy of only letting customers select lockers with available capacity \citep{amazon}. Note that this comes at the expense of our feasibility check strongly erring on the side of caution: We classify requests that can be feasibly allocated for some but not all possible realizations of pickup times as overall infeasible. There might thus be cases where, in hindsight, a request could have been delivered to the locker despite not passing the feasibility check.
    \item We do not allow parcels to be reallocated once they are assigned to a specific compartment as this would increase the operational complexity of locker deliveries.
\end{enumerate}

\subsection{Sequential Decision Problem} \label{SDP}
Leveraging the framework by \cite{powell2022}, we model the DPLDMP as a sequential decision problem.

\subsubsection{Decision Epoch.} \label{sdp_epochs} 
A decision epoch $k=1,\dots,\infty$ is triggered by one of two possible events: 
\begin{enumerate}
    \item \textit{Request arrival.} Whenever a request arrives at $t\in\mathcal{T}$, new information is revealed that necessitates a demand control decision.
    \item \textit{End of day.} At the end of each day, the provider has to allocate orders scheduled for delivery on the subsequent day. Consequently, a decision epoch arises whenever the system reaches $t=T+1$.
\end{enumerate}
Each decision epoch corresponds to a combination of day $\tau$ and point in time $t$. Note that we use index $k$ (yielding $\tau_k,t_k$) when explicitly referring to a specific decision epoch $k$. 

\subsubsection{Pre-Decision State.} \label{sdp_preds}
The pre-decision state variable $S_k$ encompasses all information necessary to make a decision in $k$ and model the system from $k$ onward:

\begin{itemize}
    \item \textit{Temporal information.} We keep track of time with the current day $\tau_k$ and point in time $t_k$. Note that including $\tau_k$ is only strictly necessary if the arrival or pickup probability distributions depend on it (e.g., the day of the week) and could otherwise be omitted. In contrast, incorporating $t_k$ is mandatory to know how much time remains until the next allocation decision.
    \item \textit{Locker occupancy.} To determine the available capacity, we require information on which locker compartments are presently occupied as well as the amount of time each parcel has already been waiting in the locker. We use the \textit{dwell time} $h\in\mathcal{H}=\{1,\dots,B-1\}$ to indicate that a parcel is spending its $h$\textsuperscript{th} day in the locker (starting with the day immediately after its allocation) with $B$ representing the maximum storage time. Note that we do not need to track parcels with a dwell time of $B$ as we know for certain that they will either be picked up or removed from the locker by the end of the current day. We represent the number of compartments of size $\delta$ that are occupied by a parcel belonging to a customer of type $c$ on the $h$\textsuperscript{th} day since its allocation with $l^k_{\delta c h}$. Overall, we can write $L_k=(l^k_{\delta c h})_{\delta\in\mathcal{D},c\in\mathcal{C},h\in\mathcal{H}}$ for the locker occupancy in $k$.
    \item \textit{Pending orders.} Besides the orders already delivered to the locker, we need to keep track of the orders scheduled for delivery in the following days, which result from previously accepted requests. We use the \textit{remaining fulfillment time} $f\in\mathcal{F}=\{1,\dots,F\}$ to indicate that an order must be assigned to a compartment in the $f$\textsuperscript{th} allocation decision from now with $f=1$ symbolizing that the order must be allocated at $T+1$ of the current day. Directly after accepting a request, its remaining fulfillment time is equal to its lead time. In other words, we use $f$ as a countdown to track the amount of time until the actual delivery, i.e., the remaining fulfillment time of each order decreases by one unit with each passing day as a part of the transition function (Section \ref{sdp_transitions}). Letting $o^k_{d c f}$ refer to the number of parcels of size $d$ belonging to a customer of type $c$ that must be allocated in the $f$\textsuperscript{th} allocation decision from now, we model the pending orders in $k$ with $O_k=(o^k_{d c f})_{d\in\mathcal{D},c\in\mathcal{C},f\in\mathcal{F}}$.
    \item \textit{Request type.} Based on the three request attributes, we construct request types $r\in\mathcal{R}=\{1,\dots,C\cdot D\cdot E\}$ with the associated customer type $c_r$, parcel size $d_r$, and lead time $e_r$. We use an artificial type $r=0$ to model the case of no request arrival. The state variable contains the type of the newly arrived request $r_k$.
\end{itemize}

In summary, we define the pre-decision state variable as $S_k=(\tau_k,t_k,L_k,O_k,r_k)$ for $k>0$ and model the set of all possible pre-decision states with the pre-decision state space $\mathcal{S}$. Additionally, we introduce the initial state $S_0$ to capture global problem parameters introduced in Section \ref{problem_description} and initial values of all parameters evolving over time \citep[][Chapter 9.4.3]{powell2022}. 

\subsubsection{Decision.} \label{sdp_decision}
The decision space $\mathcal{X}(S_k)$ encompasses all feasible decisions in $S_k$. We generally use $X_k\in\mathcal{X}(S_k)$ to refer to the decision in $k$. The specific type of decision depends on $t_k$ (Section \ref{sdp_epochs}):

\textit{Demand control.} 
If $t_k\in\mathcal{T}$, the provider must make a demand control decision regarding request $r_k$. We denote it by $g_k$ with $g_k=0$ encoding rejection and $g_k=1$ acceptance. For ease of exposition, we model rejections due to infeasibility as a special case with a reduced decision space. To determine the demand control decision space $\mathcal{X}(S_k)=\mathcal{G}(S_k)\subseteq\{0,1\}$, the provider performs a feasibility check.

To classify a newly arrived request in decision epoch $k$ as feasible, the provider must prove that it can be feasibly allocated for any pickup time realization of itself as well as the parcels represented by $O_k$ and $L_k$. Therefore, we need to check whether a tentative allocation plan spanning the next $F$ days exists where all orders in $O_k$ and the request are allocated to sufficiently large compartments without reallocation over time while maintaining the assignment of orders in $L_k$ and respecting the limited number of compartments $Q_\delta$. Assuming for the sake of the feasibility check that all customers make use of the maximum storage time, we formalize these constraints in Appendix \ref{MIP_feasibility_check}. If a feasible solution to constraints (\ref{cap1})-(\ref{int}) in Appendix \ref{MIP_feasibility_check} exists, $\mathcal{G}(S_k)=\{0,1\}$ and $\mathcal{G}(S_k)=\{0\}$ otherwise. Note that the tentative allocation plan only serves to determine the demand control decision space and is discarded afterwards.

\textit{Allocation.} 
In $t_k=T+1$, the provider must allocate parcels scheduled for delivery on the following day, i.e., all pending orders in $O_k$ with $f=1$, to available and compatible locker compartments. Let $a^k_{d \delta c}$ encode the number of parcels of size $d$ belonging to a customer of type $c$ that are allocated to a compartment of size $\delta$ (with $\delta\geq d$) in $k$ with $A_k=(a^k_{d \delta c})_{d\in\mathcal{D},\delta\in\{d,\dots,D\},c\in\mathcal{C}}$. To maintain feasibility, we require the allocation decision to comply with the constraints imposed by the feasibility check. The allocation decision space $\mathcal{X}(S_k)=\mathcal{A}(S_k)$ encompasses all feasible solutions to constraints (\ref{cap1})-(\ref{int_a}) in Appendix \ref{MIP_feasibility_check}.

\subsubsection{Post-Decision State.} \label{sdp_pds}
After making a decision in $S_k$, the system deterministically transitions into the post-decision state $S_k^x=(\tau_k,t_k,L_k^x,O_k^x)$. We model this with the transition function $S_k^x=S^{Mx}(S_k,X_k)$. In case of demand control, the locker occupancy remains unchanged ($L^x_k = L_k$). The pending orders $O_k$ are updated to $O_k^x$ with $o^{kx}_{d c f}=o^k_{d c f}+\mathbf{1}(d=d_{r_k}, c=c_{r_k},f=f_{r_k},g_k=1)$ and $\mathbf{1}(\cdot)$ symbolizing the indicator function. Each allocation decision marks the end of the current day, and we can directly update the remaining fulfillment time $f$ of all pending orders as well as the the dwell time $h$ for the locker occupancy, leading to $o^{kx}_{dcf} = o^k_{dc,f+1}$ for $f=1,\dots,F-1$ with $o^{kx}_{dcF} = 0$ and $l^{kx}_{\delta c h} = l^k_{\delta c, h-1}$ for $h>1$ with $l^{kx}_{\delta c 1} = \sum_{d=1}^\delta a^k_{d \delta c}$. We denote the post-decision state space by $\mathcal{S}^x$.

\subsubsection{Exogenous Information.} \label{sdp_transitions} 
The exogenous information $W_{k+1}=(\tau_{k+1},t_{k+1},r_{k+1},P_{k+1})\in\mathcal{W}(S_k^x)$ comprises all information that is revealed when transitioning from $S_k^x$ to $S_{k+1}$. We formalize this with the transition function $S_{k+1}=S^{MW}(S_{k}^x,W_{k+1})$ and represent the probability of observing $W_{k+1}$ when in $S_k^x$ with $\mathbb{P}(W_{k+1}|S_k^x)$. Note that while we can deterministically infer $\tau_{k+1}$ from $S_k^x$, we model it as part of $W_{k+1}$ for the sake of readability. In contrast, $t_{k+1}$ and $r_{k+1}$ depend on the stochastic request arrival process. To model the second source of uncertainty, the parcel pickup, $p_{\delta ch}^{k+1}$ denotes the number of compartments of size $\delta$ that had been occupied by a parcel belonging to a customer of type $c$ for a dwell time of $h$ in $S_k^x$ and become available again during the transition to $S_{k+1}$. In aggregated form, we can write $P_{k+1} = (p_{\delta ch}^{k+1})_{\delta\in\mathcal{D},c\in\mathcal{C},h\in\mathcal{H}}$. Starting from $S_k^x$, we determine the pending orders and locker occupancy in the next pre-decision state $S_{k+1}$ through $O_{k+1}=O_k^x$ and $l^{k+1}_{\delta ch}=l^{kx}_{\delta ch}-p^{k+1}_{\delta c h}$.

\subsubsection{Reward, Policy, and Objective Function.} \label{sdp_objective}
We define the reward function $R(S_k, X_k)$ as follows: If the provider accepts a request during demand control, the reward function takes on the value of the priority weight $m_c$ associated with the request's customer type $c_{r_k}$, i.e., $R(S_k,g_k = 1)= m_{c_{r_k}}$, and $R(S_k,g_k=0)=0$ in case of rejection. The allocation decision yields no immediate reward ($R(S_k,A_k)=0$).

The solution to a sequential decision problem is a policy $\pi\in\Pi$. The policy maps states to decisions, denoted by $X_k^\pi(S_k)$. The optimal policy $\pi^*$ maximizes the objective function. Due to the infinite planning horizon of the DPLDMP, we cannot simply formulate the objective as the expected sum of rewards over all decision epochs as it would grow to infinity. To avoid this, we could introduce a discount factor. However, from a theoretical perspective, discount factors are not well-suited for value function approximation in an infinite-horizon setting, which is an essential part of our solution strategy \citep[see][Chapter 10.4 for details]{sutton+barto}. An alternative option, which fits better to our needs, is to define the objective as the expected average reward per decision epoch with reward rate $\bar{R}^*$ \citep[][Chapter 10.3]{sutton+barto}:
\begin{equation*} \label{eq_objective}
    \bar{R}^* = \max_{\pi\in\Pi}\left\{\lim_{K\rightarrow\infty}\frac{1}{K}\sum_{k=1}^K\mathbb{E}^\pi\left[R\left(S_k,X_k^\pi\right(S_k))|S_0\right]\right\}
\end{equation*}

\subsubsection{Value Function and Opportunity Cost.} \label{sdp_valuefunc}
The value function $V(S)$ represents the value of being in a given state $S\in\mathcal{S}\cup\mathcal{S}^x$ and, in the average reward setting, equals the expected sum of rewards adjusted by $\bar{R}^*$ if we start in $S$ and apply the optimal policy $\pi^*$ from that point onward \citep{mahadevan_1996}. We can define the value function recursively with the Bellman equation \citep{sutton+barto}:
\begin{align} 
    V(S_k) & =\max_{X_k\in\mathcal{X}(S_k)}\left\{R(S_k,X_k) - \bar{R}^* + V(S_k^x)\right\} \label{value_function}\\
    &=\max_{X_k\in\mathcal{X}(S_k)}\left\{R(S_k,X_k) - \bar{R}^* + \sum_{W_{k+1}\in\mathcal{W}(S_k^x)} \mathbb{P}(W_{k+1}|S_k^x)\cdot V(S_{k+1})\right\} \label{value_function2}
\end{align}

Considering the demand control decision for a feasible request, the first expression can be reformulated to $V(S_k)=\max\left\{m_{c_{r_k}}-\bar{R}^* +V(S^{Mx}(S_k,g_k=1));-\bar{R}^*+V(S^{Mx}(S_k,g_k=0))\right\}.$ Defining $\Delta V(S_k^x)$ as the opportunity cost \citep{talluri+vanryzin}, i.e., the difference in post-decision state values caused by accepting the request in decision epoch $k$, we can rearrange the terms and observe that under the optimal policy $\pi^*$, a feasible request is accepted if and only if the following holds:
\begin{equation} \label{optimal_dc}
    m_{c_{r_k}} \geq V(S^{Mx}(S_k,g_k=0)) - V(S^{Mx}(S_k,g_k=1)) = \Delta V(S_k)
\end{equation}
In other words, the immediate reward of accepting the request, which is equal to $m_{c_{r_k}}$, must be at least as large as the request’s opportunity cost due to the expected displacement of future customers.

\textbf{Insight 1:} $\Delta V(S_k)$ \textit{is nondecreasing with increasing parcel size} (proof in Appendix \ref{monotonicity}). From this, we infer that requests with smaller parcels tend to be, ceteris paribus, more likely to get accepted under $\pi^*$ than larger parcels. Intuitively speaking, smaller parcels offer more `upgrading' possibilities and can therefore be allocated with more flexibility.

\subsection{Example} \label{sdp_example}
\begin{figure} 
\centerline{\includegraphics[scale = 1]{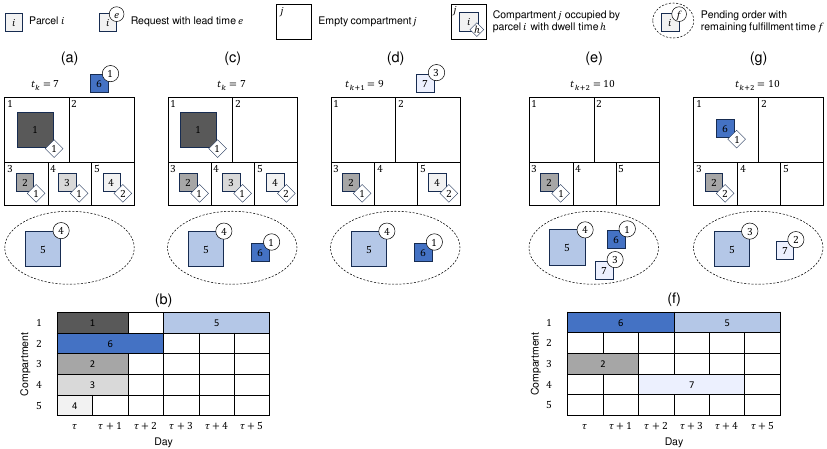}} 
\caption{Illustrative example. \label{figure_states}} 
\caption*{\footnotesize \textit{Note.} Parcel and compartment indices are for didactic purposes only. (a) Demand control pre-decision state $S_k$. (b) Tentative allocation plan in $k$. (c) Demand control post-decision state $S_k^x$. (d) Demand control pre-decision state $S_{k+1}$. (e) Allocation pre-decision state $S_{k+2}$. (f) Tentative allocation plan in $k+2$. (g) Allocation post-decision state $S^x_{k+2}$.} 
\end{figure}

To illustrate the sequential decision problem presented in the previous section, we provide a stylized example in Figure \ref{figure_states}. For clarification, we index parcels by $i$ and compartments by $j$ to explicitly refer to the respective elements in the figure. Note that we introduce these indices for didactic purposes only. 

The example includes two compartment and parcel sizes $\delta,d\in\{1,2\}$ and a parcel locker with $Q_2=2$ large ($j=1,2$) and $Q_1=3$ small compartments ($j=3,4,5$). Each day consists of $T=9$ points in time and the allocation decision in $T+1=10$. We focus on a given day $\tau$. For ease of exposition, we consider a single customer type ($C=1$) with priority weight $m_1=1$ such that $L_k$ and $O_k$ can be represented by two-dimensional matrices. The maximum storage time is $B=3$ days, and the lead time is up to $E=6$ days. For the sake of brevity, we omit request types and instead reference requests by their index $i$ in Figure \ref{figure_states}.

Figure \ref{figure_states}(a) depicts the pre-decision state $S_k$ at $t_k=7$. A new request $i=6$ with a small parcel ($d=1$) and lead time $e=1$ arrives. Two small and one large compartment are currently occupied with dwell time $h=1$ and one small compartment with $h=2$, yielding $L_k = \big(\begin{smallmatrix}
  2 & 1\\
  1 & 0
\end{smallmatrix}\big)$. Moreover, the single pending order $i=5$ is associated with a large ($d=2$) parcel and remaining fulfillment time $f=4$, which we track with $O_k=\big(\begin{smallmatrix}
  0 & 0 & 0 & 0 & 0 & 0\\
  0 & 0 & 0 & 1 & 0 & 0
\end{smallmatrix}\big)$. The pre-decision state equals $S_k=(\tau,7,\big(\begin{smallmatrix}
  2 & 1\\
  1 & 0
\end{smallmatrix}\big),\big(\begin{smallmatrix}
  0 & 0 & 0 & 0 & 0 & 0\\
  0 & 0 & 0 & 1 & 0 & 0
\end{smallmatrix}\big),6)$.

To determine the decision space, the provider performs the feasibility check, i.e., tries to prove the existence of a feasible tentative allocation plan that includes request $i=6$ and assumes all customers make use of the maximum storage time $B=3$. Figure \ref{figure_states}(b) illustrates an exemplary tentative plan where each row corresponds to a compartment $j$ and each bar to a parcel $i$. Given that we do not allow reallocation over time, parcels $i=1,2,3,4$ must remain in their respective compartments. Note that the bars representing these parcels span less than 3 days as we can infer from their dwell time that a certain portion of their maximum storage time has already elapsed. For pending order $i=5$ and request $i=6$, the allocation to a specific compartment is only preliminary and not relevant for future decision making. As we can construct a feasible tentative plan, request $i=6$ is classified as feasible, leading to the decision space $\mathcal{G}(S_k)=\{0,1\}$.

Assuming the provider accepts the request ($g_k=1$), we earn a reward of $R(S_k,g_k=1)=1$, and the system transitions into the post-decision state $S_k^x=(\tau,7,\big(\begin{smallmatrix}
  2 & 1\\
  1 & 0
\end{smallmatrix}\big),\big(\begin{smallmatrix}
  1 & 0 & 0 & 0 & 0 & 0\\
  0 & 0 & 0 & 1 & 0 & 0
\end{smallmatrix}\big))$ in Figure~\ref{figure_states}(c).

The next decision epoch $k+1$ is triggered by a request arrival $i=7$ in $t_{k+1}=9$. Apart from the new request, information on collected parcels is revealed. The pre-decision state $S_{k+1}$ in Figure \ref{figure_states}(d) shows that compartments $j=1,4$ have become available again due to the respective customers retrieving their parcels. The provider again performs the feasibility check by attempting to construct a feasible tentative allocation plan and subsequently makes the demand control decision. In the example, $i=7$ is accepted.

The sequence of demand control decisions carries on until the system reaches the end of the current day. Figure \ref{figure_states}(e) illustrates this with $S_{k+2}$ in $t_{k+2}=10$. We observe that $i=4$ was picked up during the transition from $S_{k+1}^x$ to $S_{k+2}$. The provider must allocate pending orders with $f=1$. More precisely, we must select a compartment for $i=6$ such that we do not violate the constraints of the feasibility check, e.g., $j=1$. We can confirm the feasibility of this decision in Figure \ref{figure_states}(f). Note that only the allocation of $i=6$ (and $i=2$ from a previous decision) is definitive. The assignments of $i=5,7$ are merely preliminary to prove feasibility.

Next, the system evolves into the post-decision state in Figure \ref{figure_states}(g). Order $i=6$ now occupies $j=1$. As the allocation signalizes the end of $\tau$, we update the dwell times and remaining fulfillment times in preparation for $\tau+1$, leading to $S_{k+2}^x=(\tau,10,\big(\begin{smallmatrix}
  0 & 1\\
  1 & 0
\end{smallmatrix}\big),\big(\begin{smallmatrix}
  0 & 1 & 0 & 0 & 0 & 0\\
  0 & 0 & 1 & 0 & 0 & 0
\end{smallmatrix}\big))$. The transition to the next pre-decision state marks the start of $\tau+1$, and the sequence of demand control and allocation decisions continues.

\textbf{Insight 2:} The allocation of $i=6$ to $j=1$ shows that \textit{it is not optimal to always allocate parcels to the smallest available and compatible compartment}. The reasoning behind this is similar to \cite{gönsch_steinhardt2015} and goes as follows: Observe that we already know for a fact that we must allocate $i=5$ to one of the two large compartments at the allocation decision on day $\tau+3$. Without loss of generality, assume that we are going to select compartment $j=1$. If we do not allocate $i=6$ to $j=1$ on day $\tau$, the available capacity on days $\tau,\tau+1,\tau+2$ in $j=1$ becomes worthless: Because the maximum storage time is $B=3$ days, we cannot guarantee that the compartment is going to be available again by $\tau+3$ for $i=5$ if we allocate a parcel to it in $\tau+1$ or $\tau+2$. This implies that we cannot feasibly allocate any parcel to $j=1$ on these two days. As a result, if we assign $i=6$ to $j=1$ on day $\tau$, we utilize the capacity without causing any displacement since the capacity could not be used otherwise anyway. By contrast, assigning $i=6$ to one of the other compartments could displace a future request. Therefore, it is optimal to select a larger compartment for $i=6$ than strictly necessary. 

\section{Solution Approach} \label{solution_approach}
In this section, we present our solution approach to obtain a policy for the DPLDMP. First, we outline our framework and the core ideas behind it in Section \ref{s_outline}. Second, we specify the cost function approximation (CFA) that governs allocation decisions in Section \ref{CFA}. Third, Section \ref{VFA} is dedicated to the parametric value function approximation (VFA) responsible for demand control.

\subsection{Outline and Motivation} \label{s_outline}
The well-known `curses of dimensionality' \citep{powell2022} render the computation of $\pi^*$ intractable for realistic problem sizes. While $\mathcal{G}(S_k)$ is manageable because $|\mathcal{G}(S_k)|\leq 2$, $\mathcal{A}(S_k)$ corresponds to a multidimensional resource allocation decision space that grows combinatorially in its dimensions. Similarly, the multidimensional matrices in $S_k,S_k^x$ and $W_k$ lead to prohibitively large state and outcome spaces $\mathcal{S},\mathcal{S}^x,\mathcal{W}(S_k)$ for real-world applications. To overcome these challenges, we need to develop a suitable solution approach. 

Apart from the two sources of uncertainty induced by request arrivals and parcel pickups, a key characteristic of the DPLDMP stems from the two types of strongly interrelated decisions arising at different points in time. Consequently, the solution framework must not only encompass tailored components to handle each decision type on its own, but also properly address their interdependencies:
\begin{itemize}
    \item Well-performing allocation decisions ensure that the locker's limited capacity is utilized efficiently and simultaneously create favorable conditions for subsequent demand control. This requires anticipating allocation decisions for pending orders and predicting new orders resulting from future demand control.
    \item As shown in (\ref{optimal_dc}), optimal demand control essentially trades off a request's immediate reward with its opportunity cost. As a result, demand control should factor in allocation decisions as they influence the available capacity in the compartments, shape the decision space for future demand control, and affect the capacity consumption caused by accepting the request, which is all reflected in the opportunity cost.
\end{itemize}

To handle the interdependencies between the two decision types, we propose a hierarchical approach. More precisely, our solution framework comprises two components, one for demand control and one for allocation, with different levels of sophistication. Given that demand control determines which requests turn into orders and aims at reserving capacity for higher-priority customers, it presents itself as a more promising lever to improve overall performance. By comparison, allocation merely serves as a secondary decision. Intuitively speaking, it is harder to compensate bad demand control decisions with good allocation decisions than vice versa. Bearing this in mind, we consciously limit the computational effort of the allocation component in favor of an elaborate demand control component.

Drawing on the framework by \cite{powell2022}, we design the individual components as follows:
\begin{itemize}
    \item Allocation is characterized by its extensive decision space. In light of this, we employ cost function approximation (CFA), which yields a parameterized deterministic optimization model. This allows us to efficiently search the decision space through standard solvers.
    \item By contrast, the demand control decision space is limited to at most two options, namely acceptance or rejection of the current request. If the provider accepts a feasible request, it turns into an order and typically stays in the system for several days depending on its lead time and the pickup time. To properly estimate the downstream impact of the decision, we apply a parametric value function approximation (VFA).
\end{itemize}
\begin{figure} 
\centerline{\includegraphics[scale = 0.9]{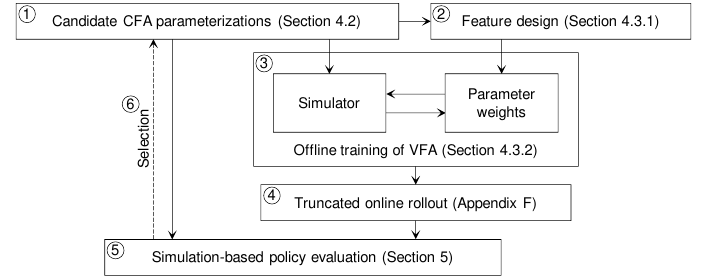}} 
\caption{Solution procedure. \label{figure_s_framework}} 
{} 
\end{figure}
Harnessing this architecture, we devise the procedure in Figure \ref{figure_s_framework} to obtain the CFA and VFA parameters:
\begin{enumerate}
    \item The vast set of potential CFA parameterizations necessitates tuning. However, the performance of a specific CFA parameter configuration also depends on demand control, whose VFA parameter weights in turn hinge on how allocation decisions are made as a result of the CFA parameters. Consequently, we need to train the VFA for each CFA parameter configuration we want to examine, making the evaluation of CFA parameters computationally expensive. To mitigate this, we predefine a promising subset of candidate parameterizations through domain knowledge. With these, we aim at creating favorable conditions for future demand control and thus reflect the influence of demand control on allocation.
    \item We embed the allocation scheme of a given CFA parameter configuration into the VFA feature design, thereby incorporating the impact of allocation into demand control. This yields a vector of features $\bm{\phi}(S_k^x)$. 
    \item To train the VFA, we have to learn the parameter weights for the features through offline simulation. The simulator requires a given CFA parameterization, which determines the simulated allocation decisions (and the feature design), as an input and returns a vector of trained weights $\bm{\theta}$.
    \item As an optional enhancement, we can combine the trained VFA with a truncated online rollout (Appendix \ref{a_rollout}). This part can be omitted in case of limited computational resources.
    \item A specific CFA parameterization together with the correspondingly trained VFA (and truncated rollout) form a candidate policy, whose performance is evaluated through simulation. 
    \item Among the evaluated candidate policies, we choose the best-performing one as the final policy.
\end{enumerate}

In summary, we take the interdependencies between the two decision types into account by contriving candidate CFA parameterizations that aim at facilitating future demand control and, conversely, by constructing the features and training the VFA for the allocation schemes induced by the CFA parameters.

\begin{table}
\caption{Tentative allocation plans.\label{overview_tentative_plans}}
\centering
\resizebox{\textwidth}{!}{\renewcommand{\arraystretch}{1.2}
\begin{tabular}{>{\raggedright}m{3.5cm} >{\centering\arraybackslash}p{4cm}
>{\centering\arraybackslash}p{2cm}
>{\centering\arraybackslash}p{3cm}
>{\centering\arraybackslash}p{4cm}>{\centering\arraybackslash}p{4cm}>{\centering\arraybackslash}p{4cm}} \hline
& \multirow{2}{*}{Pickups} & \multirow{2}{*}{Discarded}
& \multicolumn{2}{c}{Objective} & \multicolumn{2}{c}{Constraints} \\
& &
& Model & Solution framework & Model & Solution framework \\
\hline
Feasibility check & Maximum storage time & Entirely 
& - & - & (\ref{cap1})-(\ref{int}), Appendix \ref{MIP_feasibility_check} & (\ref{cap1})-(\ref{int}), Appendix \ref{MIP_feasibility_check}  \\
Allocation decision & Maximum storage time & Partly 
& (\ref{value_function}) & (\ref{obj_s_first})/(\ref{obj_w_first}), Appendix \ref{CFA_details} & (\ref{cap1})-(\ref{int_a}), Appendix \ref{MIP_feasibility_check} & (\ref{CFA_cap1})-(\ref{CFA_int}), Appendix \ref{CFA_details} \\
VFA features & Sampled & Entirely 
& - & (\ref{vfa_s_first})/(\ref{vfa_w_first}), Appendix \ref{VFA_details} & - & (\ref{VFA_cap1})-(\ref{VFA_int}), Appendix \ref{VFA_details} \\
\hline
\end{tabular}}
\end{table}

Throughout both our model (Section \ref{SDP}) and our solution framework, tentative allocation plans play a crucial role. For orientation, we provide a corresponding overview in Table \ref{overview_tentative_plans}. Basically, tentative allocation plans are relevant in three different contexts: for the feasibility check, allocation decision, and VFA features. In all three cases, we need to make an assumption on when orders are picked up to construct the plans. For the feasibility check and allocation decision, we use the maximum storage time. To include information on the stochastic pickup behavior in our VFA features, we generate sampled pickups. Note that the tentative plans for the feasibility check and VFA features get fully discarded immediately after extracting the relevant information. By contrast, from the tentative plan for the allocation decision at the end of each day, the decisions that refer to the current day are implemented and only the remaining parts of the plan get discarded. From a theoretical perspective, the feasibility check corresponds to a constraint satisfaction problem whereas the tentative plans for the allocation decision and VFA features correspond to optimization problems. In our solution framework, we solve the resulting instances with standard software.

\subsection{Cost Function Approximation} \label{CFA}
Cost function approximation (CFA) involves solving a parameterized deterministic optimization model where the objective or constraints are modified to induce decisions that perform well over time and under uncertainty. For the problem at hand, we apply this concept to the allocation decision and tweak the immediate reward such that it provokes decisions conducive to demand control. This entails two steps: First, designing the parameterization (Section \ref{CFA_design}), and second, determining the parameter values (Section \ref{CFA_tuning}).

\subsubsection{Design of Parameterization.} \label{CFA_design}
As a starting point for modifying the immediate reward function of allocation decisions, we identify two favorable circumstances for subsequent demand control:
\begin{enumerate}
    \item In general, larger compartments tend to be more valuable as they fit more parcel sizes and offer more flexibility. Although the example in Section \ref{sdp_example} proves that it is not optimal to always assign parcels to the smallest available compartment, this principle still holds in many cases.
    \item Given the assumptions of the feasibility check, having a specific compartment available for multiple consecutive days increases the likelihood of incoming requests being feasible. The allocation decision is based on essentially the same constraints as the feasibility check. In light of this, we aspire to allocate orders such that we conserve available capacity in the compartments across several days to facilitate the construction of feasible tentative plans in the future. 
\end{enumerate}

Bearing these two considerations in mind, we introduce the concept of \textit{capacity windows}. With this term, we refer to a stretch of available capacity in the tentative allocation plan. It is characterized by its length $\lambda$, i.e., the number of allocation decisions (days) it spans across, and the size $\delta$ of the compartment in which it arises. For a specific tentative allocation plan, we let $w_{\delta\lambda}$ denote the number of capacity windows in compartments of size $\delta$ with length $\lambda$. As an example, the tentative allocation plan in Figure \ref{figure_states}(b) contains one capacity window of length five ($w_{15}=1$) and two windows of length four ($w_{14}=2)$ in the small compartments as well as one window of length one and one of length three in the large compartments ($w_{21}=w_{23}=1$). Capacity windows serve as the basis for our CFA design as they encapsulate both the notion of preserving capacity across multiple days as well as the relevance of the compartment size. 

In the objective function, we weight $w_{\delta\lambda}$ with the corresponding CFA parameters represented by $v_{\delta\lambda}$. Since the allocation decision itself yields no immediate reward, we obtain the following CFA objective: 
\begin{equation}
    \max \sum_{\delta\in\mathcal{D}}\sum_{\lambda\in\mathcal{F}} v_{\delta\lambda}w_{\delta\lambda} \label{cfa_obj}
\end{equation}
To properly link $w_{\delta\lambda}$ to the allocation decision $A_k$ and the resulting tentative allocation plan, we formalize the CFA decision space with constraints (\ref{CFA_cap1})-(\ref{CFA_int}) in Appendix \ref{CFA_details}.

\subsubsection{Parameter Selection.} \label{CFA_tuning} 
After specifying the CFA design, we need to determine suitable values for $v_{\delta\lambda}$. A key challenge is posed by the large space of potential parameterizations that grows combinatorially with the number of compartment sizes $D$ and window lengths $F$. To tackle it, one would typically resort to methods from the realm of stochastic search \citep[][Chapter 11.12]{powell2022}. However, these methods require evaluating each generated parameter configuration. In our case, this is computationally expensive as the CFA parameters are also embedded in the VFA feature design and part of the VFA training procedure. More specifically, we have to retrain the VFA whenever the CFA parameters are adapted and then measure the performance of the resulting policy through simulation. Consequently, evaluating a sizeable number of parameters is hardly tractable. Instead, we limit the number of candidate parameterizations a priori through domain knowledge, thereby sidestepping the issue of excessive computational effort for parameter tuning.

Essentially, capacity windows are characterized by their length $\lambda$ and the compartment size~$\delta$. Accordingly, we propose two parameter configurations, each geared towards one of these attributes:
\begin{enumerate}
    \item Since larger compartments tend to offer more flexibility, we seek to conserve available capacity in them and preferably allocate parcels to smaller compartments. In other words, we prioritize available capacity in decreasing hierarchical order of the compartment size.
    \item Gaps of available capacity that stretch across multiple allocation decisions in the tentative plan are favorable for demand control. To operationalize this, we maximize the number of capacity windows in decreasing hierarchical order of their length.
\end{enumerate}

For the specific parameterizations, we refer to Appendix \ref{CFA_details}. Each of these two parameter configurations addresses one of the two attributes of capacity windows, compartment size and length, individually. To cover both dimensions simultaneously, we can optimize the objectives in lexicographic order. In total, the CFA is specified by the lexicographic order in which the objectives are optimized and its decision space formalized by (\ref{CFA_cap1})-(\ref{CFA_int}) in Appendix \ref{CFA_details}. We solve the resulting integer program (IP) with standard software, thereby handling the curse of dimensionality in the allocation decision space.

\subsection{Value Function Approximation} \label{VFA}
If an oracle were to provide us the optimal value function $V(S)$, applying the decision criterion in (\ref{optimal_dc}) becomes trivial. Essentially, it boils down to trading off the request's immediate reward, i.e., its priority weight, with the opportunity cost. The opportunity cost captures the potential displacement of future customers: Each accepted request consumes capacity, which might force the provider to reject future requests.

As laid out in Section \ref{s_outline}, the curses of dimensionality render the computation of $V(S)$ intractable for realistic problem sizes, compelling us to rely on approximation. Specifically, we employ a parametric value function approximation (VFA) to overcome the curse of dimensionality in the state space. Instead of computing the value for each state individually, we learn a parametric function with a vector of weights $\bm{\theta}$ to obtain value estimates $\hat{V}(S|\bm{\theta})$. As an input for the VFA, we devise a set of features $\bm{\phi}(S_k^x)$ to extract relevant information from the state variable and combine them in a suitable functional form, which is detailed in Section \ref{VFA_features}. Subsequently, Section \ref{VFA_training} sheds light on the procedure to learn the parameter weights $\bm{\theta}$. 

\subsubsection{Feature Design.} \label{VFA_features}
As a preliminary consideration, we first establish which values we seek to approximate given that the value function can be computed for pre-decision states $S_k$, post-decision states $S_k^x$ or pairs of pre-decision states and decisions $(S_k,X_k)$. The latter corresponds to the `Q-values' typically used in RL, where rewards are modeled as a random variable. By contrast, we want to exploit the fact that rewards are a deterministic function of $S_k$ and $X_k$ in the DPLDMP, which is not possible with Q-values. If we were to approximate the values for $S_k$, we would have to explicitly calculate the expectation in (\ref{value_function2}), which is not tractable because of the curse of dimensionality in the outcome space. To sidestep this, the best-suited choice is to approximate the values of the post-decision states $S_k^x$.

In a next step, we need to identify an adequate representation of the information contained in $S_k^x$ in the form of a vector of features $\bm{\phi}(S^x_k)$. In theory, one could directly use $S^x_k$ itself. However, this comes at the disadvantage of a relatively large number of features and hence parameters to learn while simultaneously foregoing the opportunity to incorporate domain knowledge. Specifically, $S^x_k$ offers no immediate insight into the implicit resource allocation task at the end of each day and the allocation logic induced by the CFA. 

Intuitively speaking, the expected future rewards and thus the value of a state hinge on the available capacity and the incoming future demand, which are both uncertain in the DPLDMP. As an underlying strategy, we aim to capture an estimate of the available capacity through our features and then evaluate what can be achieved with it in terms of expected future rewards through the parameter weights $\bm{\theta}$. This general idea has proven successful in other domains, such as dynamic vehicle routing \citep[e.g.,][]{ulmer_etal_2020} or integrated demand management and vehicle routing problems \citep[e.g.,][]{koch+klein_2020}. 

To measure available capacity, we revert to the concept of capacity windows introduced in Section \ref{CFA_design}, i.e., stretches of available capacity in a tentative allocation plan that are characterized by their length $\lambda$ and compartment size $\delta$. For the CFA, capacity windows are determined based on the allocation decision space, which assumes that all customers use the maximum storage time. For the VFA features, we slightly adapt this to get an estimate of available capacity that also reflects the uncertain pickup behavior of customers.

As a basis for the features, we first generate a total of $U$ scenarios with sampled pickups for all pending orders $O_k^x$ and orders currently in the locker $L_k^x$. Note that in the latter case, we can exploit the information revealed by the dwell time $h$ and the current point in time $t_k$ in the form of conditional pickup probabilities. For each scenario $u=1,\dots,U$, we then construct a tentative allocation plan where all orders in $O_k^x$ are preliminarily allocated to compartments. As in Section \ref{problem_statement}, we do not allow reallocation of orders over time and must respect the locker's limited capacity. We formalize the construction of the tentative plan with an IP in Appendix \ref{VFA_details}. Note that the tentative plan with sampled pickups covers a slightly longer planning horizon $\hat{\mathcal{F}}=\{1,\dots,F+B-1\}$ to fully include the pickup times of orders allocated in $f=F$.

To incorporate the allocation scheme in the feature design, we use the CFA parameters $v_{\delta\lambda}$ as objective function coefficients for the tentative plans. Note that we do not aim to perfectly predict future allocation decisions; the main purpose is to get an estimate on available capacity with reasonable computational effort that allows a differentiated evaluation according to compartment size as well as `connectedness' across multiple days. Given that we need to create a tentative plan for each scenario and that $U$ should be sufficiently large to obtain a reliable estimate, we reduce the computational burden by consciously neglecting the dynamics of pickups and constructing the tentative plans as if all pickup times were known upfront.

In each scenario $u$, we extract the number of capacity windows $\hat{w}^u_{\delta\lambda}$ per compartment size $\delta$ and length $\lambda$ from the tentative allocation plan. Averaging across all scenarios, we obtain $\hat{w}_{\delta\lambda}=\frac{1}{U}\sum_{u=1}^U\hat{w}^u_{\delta\lambda}$ as input features for the VFA. Regarding the functional form, we opt for a linear approximation architecture with parameter weights $\theta_{\delta\lambda}$ and a constant intercept $\theta_0$. Note that including an intercept is not only strongly recommended in general \citep[][Chapter 4.4]{kutner_etal_2004}, but also necessary in our specific setting: Even if all features $\hat{w}_{\delta\lambda}$ take on the value zero, meaning that we estimate to have no available capacity during the limited planning horizon of the tentative allocation plan, we still expect to generate rewards in the long run once the capacity becomes unoccupied again, which is captured by the intercept.

Letting $\bm{\phi}(S_k^x)^T=(1,\hat{w}_{11},\dots,\hat{w}_{1,F+B-1},\dots,\hat{w}_{D,F+B-1})$ denote the vector of features and $\bm{\theta}^T=(\theta_0,\theta_{11},\dots,\theta_{1,F+B-1},\dots,\theta_{D,F+B-1})$ the vector of parameter weights, we specify $\hat{V}(S|\bm{\theta})$ as follows:
\begin{equation} \label{VFA_architecture}
    \hat{V}(S|\bm{\theta}) = \bm{\theta}^T\bm{\phi}(S_k^x)=\theta_0 + \sum_{\delta\in\mathcal{D}}\sum_{\lambda\in\hat{\mathcal{F}}} \theta_{\delta\lambda}\hat{w}_{\delta\lambda}
\end{equation}
On the surface, (\ref{VFA_architecture}) bears a strong resemblance to (\ref{cfa_obj}). However, note that they differ in two critical ways:
\begin{enumerate}
    \item \textit{Calculation of capacity windows.} For the CFA, $w_{\delta\lambda}$ is calculated assuming every customer makes use of the maximum storage time, whereas the VFA features $\hat{w}_{\delta\lambda}$ result from averaging across multiple sample realizations of pickups. We refrain from incorporating sampled pickups in the CFA for two reasons: First, while we can compute each scenario separately for the VFA, we would have to jointly optimize the allocation decision over all scenarios for the CFA. This would lead to a considerable increase in the computational effort. Second, the worst-case scenario is highly relevant for subsequent demand control as it is the foundation for the feasibility check, which motivates gearing allocation decisions towards it.
    \item \textit{Determination of parameters.} The VFA parameter weights $\bm{\theta}$ aim at approximating the value function of a policy and are the result of a simulation-based learning process that requires a CFA parameter configuration as an input (Sections \ref{s_outline} and \ref{VFA_training}). By contrast, the CFA parameters $v_{\delta\lambda}$ are the result of tuning (Section \ref{CFA_tuning} and \citealp[][p. 613]{powell2022}). Note that we cannot simply use the trained weights $\bm{\theta}$ in the CFA objective because they are always trained for a specific allocation scheme. If we change the CFA parameters, we must retrain $\bm{\theta}$ (and adapt the objective for computing the tentative allocation plans for $\bm{\phi}(S_k^x)$).
\end{enumerate}

\subsubsection{Training.} \label{VFA_training}
To train the VFA parameter weights $\bm{\theta}$, we combine different RL techniques in a simulation-based learning process that is executed offline. Our algorithmic procedure encompasses two types of updates: Upon each simulated demand control decision, we update the parameter weights incrementally based on temporal difference (TD) learning \citep{sutton+barto}. At the end of selected days, we additionally perform a second type of update using experience replay \citep[ER,][]{lin_1993}. For the ER updates, we randomly sample a batch of previously encountered demand control decision epochs from a replay memory and apply ridge regression. The purpose of this step is to increase data efficiency and stabilize the learning process. Moreover, it serves as an opportunity to embed domain knowledge.

In the subsequent paragraphs, we delve into the algorithmic procedure summarized in Algorithm~\ref{alg_VFA_training_short}. As a first step, we briefly introduce the hyperparameters: The simulation runs for a total of $\tau^\text{max}$ days and we use $n$ to refer to the simulated decision epochs. For decision making, we introduce an exploration scheme $\varepsilon_n$. The ER parameters specify the size $\kappa$ of the replay memory $\mathbb{M}$ and the size $\chi$ of the samples drawn from it as well as the start $\eta$ and frequency $\zeta$ of ER updates. Lastly, we require a prespecified CFA parameterization $v_{\delta\lambda}$ as it forms the foundation of the VFA features $\bm{\phi}(S_k^x)$ and governs allocation decisions during the simulation. We initialize the algorithm by setting all parameter weights $\bm{\theta}$ to zero. Furthermore, the algorithm starts with an empty replay memory $\mathbb{M}=\{\}$ and in decision epoch $n=0$ with an initial post-decision state $S_0^x$. We provide a more detailed pseudocode along with the full set of hyperparameters in Appendix~\ref{VFA_details}.

\begin{algorithm}
\fontsize{10}{14}\selectfont
\caption{Training of VFA parameter weights $\bm{\theta}$}\label{alg_VFA_training_short}
    \textbf{Initialization:} $\bm{\theta}=\mathbf{0},\mathbb{M}=\{\},n=0, S_0^x$
\begin{algorithmic}[1]
\For{$\tau=1,\dots,\tau^{\text{max}}$}
    \Repeat
        \State $n \gets n+1$
        \State sample exogenous information $W_n$ to obtain $S_n = S^{MW}(S^x_{n-1},W_n)$
        \If{$t_n\leq T$}
            \State make $\varepsilon$-greedy demand control decision $g_n$ to obtain $S_n^x=S^{Mx}(S_n,g_n)$
            \State perform TD update and add experience to $\mathbb{M}$
            \If{$|\mathbb{M}|>\kappa$}
                \State remove oldest experience from $\mathbb{M}$
            \EndIf
        \EndIf
    \Until{$t_n=T+1$}
    \State make allocation decision $A_n$ to obtain $S_n^x=S^{Mx}(S_n,A_n)$
    \If{$(\tau_n>\eta)\wedge(\tau_n\bmod\zeta=0)$}
        \State perform ER update \label{ER_short}
    \EndIf
\EndFor
\end{algorithmic}
\end{algorithm}

During each simulated day, we generate the next decision epoch by sampling exogenous information $W_n$ and transitioning into a new pre-decision state $S_n$. In case of a request arrival, we make an $\varepsilon$-greedy demand control decision, i.e., we choose the decision that is deemed optimal according to our current parameter weights with probability $1-\varepsilon_n$ and select a random decision out of $\mathcal{G}(S_n)$ with probability $\varepsilon_n$. This induces the algorithm to try decisions that are perceived suboptimal according to the current parameter weights to verify their evaluation and ensure sufficient exploration. After demand control, the system evolves into $S_n^x$. Based on this transition, we compute the TD error to update $\bm{\theta}$. The data collected during decision epoch $n$ is stored as an experience. If the size of the replay memory exceeds $\kappa$, we delete the oldest experience.

The sequence of demand control decisions and TD updates carries on until the day concludes with the allocation decision in $T+1$, which is shaped by the CFA parameterization $v_{\delta\lambda}$. In addition, after the first $\eta$ days, we perform an ER-based update for $\bm{\theta}$ every $\zeta$ days. The threshold $\eta$ serves to ensure that a sufficient amount of experience has accumulated in the replay memory. For the ER update, we generate a simple random sample without replacement from $\mathbb{M}$. The sample contains $\chi$ previously encountered decision epochs , each associated with a stored experience. Based on this data, we can compute update targets. Essentially, for each sampled decision epoch, we determine what demand control decision we would make according to our current parameter weights. In other words, we replay past experiences and determine how we would decide based on our current knowledge if we were to encounter a previous state again. From this, we construct a value estimate that serves as the prediction target in a ridge regression to obtain updated parameter weights.

Note that our version of ER differs from its standard implementation in RL in the following four ways:
\begin{enumerate}
    \item \textit{Post-decision state values.} Instead of Q-values, we approximate post-decision state values. This entails adaptations that make the ER updates less off-policy, i.e., the difference between the behavior policy used to generate data and the target policy to be approximated is less pronounced \citep[cf.][]{mnih_etal_2015}. 
    \item \textit{Ridge regression.} During the ER update, we perform ridge regression to hedge against multicollinearity issues. To illustrate, if the number of longer capacity windows reduces after a request acceptance, this is accompanied by simultaneous shifts in other features, i.e., the number of shorter windows increasing. Multicollinearity does not generally inhibit the predictive capabilities of the regression as long as new observations lie in the region of previous data \citep[][]{kutner_etal_2004}. However, given that the region of frequently encountered states changes over the course of training, we cannot guarantee that this prerequisite holds.
    \item \textit{Two types of updates.} Our approach of regular TD and periodic ER updates extends the `Combined-Q' algorithm by \cite{zhang+sutton2017} that uses a combination of TD and ER updates in every decision epoch. Our reasoning for less frequent ER updates is as follows: The TD update is based on a single, brand new observation, i.e., the data of the current decision epoch. As a complement, ER draws on a batch of previously stored experiences and keeps the value estimates in check for a larger sample of the state space, thereby increasing data efficiency and stabilizing the learning process \citep{mnih_etal_2015}. However, focusing too much on ER runs the risk of fitting the value function to a large portion of states visited under previous policies that are in fact not relevant for the optimal policy. In VFA, we cannot hope to approximate every state value perfectly. Consequently, we mainly base the learning process on `fresh' data through TD updates and merely support it with less frequent ER updates. 
    \item \textit{Enforcing structure.} Additionally, the ER update can be used to enforce structural properties in the value function by imposing constraints on the parameter weights (Section \ref{comp_study} and Appendix \ref{policy_details}).
\end{enumerate}

\section{Computational Study} \label{comp_study}
In the following, we present our computational study. After explaining its setup in Section \ref{cs_design}, the subsequent analysis of our experiments serves two purposes: Firstly, we aim to assess the overall performance of our solution framework and gain an understanding of the contribution of individual algorithmic components (Section \ref{cs_performance}). Secondly, we aspire to provide managerial insights (Section \ref{cs_managerial}). 

\subsection{Design} \label{cs_design}
Our study encompasses nine \textit{settings} that differ with regard to the customer prioritization scheme and pickup behavior. In this section, we introduce the corresponding parameters (Sections \ref{cs_inv_params} and \ref{cs_dep_params}) and the investigated policies (Section \ref{cs_policies}). Furthermore, we define the metrics used for evaluation (Section \ref{cs_kpi}). 

\subsubsection{Setting-invariant Parameters.} \label{cs_inv_params}
The following parameters apply to all settings:

\textit{Resources.} We include three compartment sizes $\mathcal{D}=\{1,2,3\}$ with $Q_1=15$, $Q_2=10$ and $Q_3=5$. For readability, we encode the compartment sizes as S, M, L for $\delta=1,2,3$ respectively.

\textit{Time.} Each day consists of $T=20$ discrete points in time with the allocation decision in $T+1=21$.

\textit{Requests.} In all settings, there are two customer types $\mathcal{C}=\{1,2\}$. At each point in time $t$, a customer of type $c=1$ arrives with probability 0.3, of type $c=2$ with probability 0.6, and with probability 0.1 there is no customer arrival. We consider $c=1$ to be \textit{premium} customers with expedited shipping, leading to a lead time of $e=1$ with probability 1. For \textit{standard} customers ($c=2$), the lead time $e$ equals 1, 2, 3, 4, or 5 days with a respective probability of 0.2, 0.2, 0.3, 0.2, and 0.1. The probabilities for the parcel sizes are identical for both customer types and are proportional to the number of compartments per size ($\frac{1}{2}$, $\frac{1}{3}$, $\frac{1}{6}$ for S, M, L).

\textit{Parcel pickups.} The maximum storage time is $B=3$ days.

\subsubsection{Setting-dependent Parameters.} \label{cs_dep_params}
The settings result from combinations of different customer priorities and pickup probability distributions. Regarding the former, we vary the priority weight of premium customers $m_1\in\{1,2,3\}$ while keeping the weight of standard customers constant ($m_2=1$). This allows us to gauge the impact of the priority weights, which is a central input parameter to the DPLDMP. 

For parcel pickups, we examine three probability distributions: With the first one, premium and standard customers exhibit the same pickup behavior (id for `\textbf{id}entical'). Under the second one, premium customers collect their parcels faster (pf for `\textbf{p}remium \textbf{f}ast'), which is motivated by empirical evidence \citep{sethuraman_etal}. Lastly, with the third distribution, we make this difference in pickup speed even more pronounced (pu for `\textbf{p}remium \textbf{u}ltrafast'). Remember that we express the pickup time as a tuple $\psi=(b,q)$ where $b$ represents the number of days after allocation and $q$ the point in time (Section \ref{problem_description}). We state the probabilities for $b$ in Table \ref{pickup_probs}. To make our settings realistic, we select the probabilities for id based on the fact that about 60\% of parcels are picked up within one day after delivery in practice \citep{pickup_norway}. For comparability, we specify the probabilities for pf and pu such that the aggregated distribution over the entire customer population remains unchanged, i.e., in each setting, 60\% of all parcels are picked up within one day, 20\% on the second day, and 20\% on the third day. We assume a uniform distribution for $q$, leading to a probability of $\frac{1}{T}=\frac{1}{20}$ for each point in time $t\in\mathcal{T}$ in all settings.

\begin{table}%{r}{4.5cm}
\centering
\renewcommand{\arraystretch}{1.2}
\parbox[t]{.40\linewidth}{
\centering
\caption{Pickup probabilities.} \label{pickup_probs}
\resizebox{0.35\textwidth}{!}{\begin{tabular}{>{\centering\arraybackslash}p{0.5cm} >{\centering\arraybackslash}p{0.5cm} >{\centering\arraybackslash}p{1.5cm}>{\centering\arraybackslash}p{1.5cm}>{\centering\arraybackslash}p{1.5cm}} \hline
 & & \multicolumn{3}{c}{Number of days after allocation $b$} \\ \cline{3-5}
 & $c$ & 1 & 2 & 3 \\ \hline
\multirow{2}{*}{id} & 1 & 0.6 & 0.2 & 0.2 \\
& 2 & 0.6 & 0.2 & 0.2 \\ \hline
\multirow{2}{*}{pf} & 1 & 0.8 & 0.1 & 0.1 \\
& 2 & 0.5 & 0.25 & 0.25 \\ \hline
\multirow{2}{*}{pu} & 1 & 0.94 & 0.04 & 0.02 \\
& 2 & 0.43 & 0.28 & 0.29 \\
\hline
\end{tabular}}}
\parbox[t]{.40\linewidth}{
\centering
\caption{Settings.} \label{settings}
\resizebox{0.35\textwidth}{!}{\begin{tabular}{>{\centering\arraybackslash}p{0.5cm} >{\centering\arraybackslash}p{0.7cm} >{\centering\arraybackslash}p{1.5cm}>{\centering\arraybackslash}p{1.5cm}>{\centering\arraybackslash}p{1.5cm}} \hline
 & & \multicolumn{3}{c}{Pickup probability distribution} \\ \cline{3-5}
 &  & id & pf & pu \\ \hline
 & 1 & 1id & 1pf & 1pu \\
$m_1$ & 2 & 2id & 2pf & 2pu \\
 & 3 & 3id & 3pf & 3pu \\
\hline
\end{tabular}}}
\end{table}

We consider all $3\cdot3=9$ combinations of prioritizations and pickup distributions. As shown in Table \ref{settings}, we encode the settings by concatenating $m_1$ with the respective abbreviation for the pickup distribution.

\subsubsection{Policies.} \label{cs_policies}
Because of the two types of decisions present in the DPLDMP, each of our policies consists of two components. For the allocation component, we test three CFA parameterizations:
\begin{itemize}
    \item \textbf{DL} (`first \textbf{d}imension, then \textbf{l}ength') is based on the concept of capacity windows (Section \ref{CFA}) and uses compartment size as the primary and window length as the secondary objective.
    \item \textbf{LD} (`first \textbf{l}ength, then \textbf{d}imension') only differs from DL in the lexicographic order of the objectives, i.e., window length serves as the primary and compartment size as the secondary objective.
    \item \textbf{BU} (`\textbf{b}ottom-\textbf{u}p') acts as a benchmark that does not rely on capacity windows and instead focuses fully on the next allocation decision. It prioritizes using smaller compartments first (details in Appendix \ref{policy_details}).
\end{itemize}

For demand control, we employ the approach presented in Section \ref{VFA} (and Appendix \ref{a_rollout} for the rollout). To assess the contribution of algorithmic modules to the overall performance, we generate three variations of the VFA by adapting or removing certain parts of Algorithm \ref{alg_VFA_training_short} in the style of an ablation study. We test each of the trained VFAs with (\textbf{RV}) and without (\textbf{V}) a rollout as well as the rollout on its own (\textbf{R}) and include a practice-inspired and a myopic benchmark. In total, we obtain nine demand control approaches:
\begin{itemize}
    \item \textbf{(R)V-CER} (`\textbf{c}onstrained \textbf{e}xperience \textbf{r}eplay') imposes constraints on the ridge regression for ER that enforce structural properties of the value function (Appendix \ref{VFA_details}).
    \item \textbf{(R)V-ER} (`\textbf{e}xperience \textbf{r}eplay') equals (R)V-CER without constraints during the ridge regression.
    \item \textbf{(R)V-TD} (`\textbf{t}emporal \textbf{d}ifference') only uses TD updates, i.e., we eliminate line \ref{ER_short} in Algorithm \ref{alg_VFA_training_short}. 
    \item \textbf{R} (`\textbf{r}ollout') solely relies on the truncated online rollout without VFA.
    \item \textbf{DLP} (`\textbf{d}eterministic \textbf{l}inear \textbf{p}rogram') is an established method (Section \ref{related_work}) and similarly applied in industry \citep{sethuraman_etal}. We adapt this concept to the DPLDMP (details in Appendix \ref{policy_details}).
    \item \textbf{FC} (`\textbf{f}easibility \textbf{c}ontrol') is a purely myopic approach that accepts every feasible request.
\end{itemize}

We investigate all combinations of the allocation and demand control approaches listed above, resulting in a total of $3\cdot9=27$ policies. To refer to the policies, we use $\pi$ with the respective demand control abbreviation in superscript and allocation in subscript, e.g., $\pi^\text{RV-CER}_\text{DL}$. 

Details on hyperparameters and how we generated instances are provided in Appendix \ref{cs_inst_generation}. We implemented all algorithms in Python 3.8 and used Gurobi 10.0.3 to solve the IPs for the feasibility check, CFA, and VFA features as well as the quadratic program for ER. The experiments were conducted on a server with two Intel Xeon E7-8890 v3 processors (2.5 GHz, 18 cores) and 512 GB RAM. On average, the feasibility check took 0.01 seconds and the allocation decision 0.03 seconds. For demand control, decision times depend on the amount of online computations and range from instantaneous (FC) to up to 3 seconds (RV). 

\subsubsection{Metrics.} \label{cs_kpi}
We use two metrics to judge the solution quality and derive managerial insights:
\begin{itemize}
    \item \textit{Objective improvement.} We compute the relative improvement in the number of accepted customers weighted by their priority compared to $\pi^\text{FC}_\text{DL}$, which is the best-performing policy with FC over all settings.
    \item \textit{Acceptance rate.} The acceptance rate equals the number of accepted requests (with a specific combination of attributes) divided by the total number of requests (with the same attribute values).
    
We calculate each metric for 30 instances (spanning 30 days each) of every setting and then compute the average per setting. For aggregate results, we subsequently compute the average over all settings.
\end{itemize}

\subsection{Performance and Contribution of Algorithmic Components} \label{cs_performance}
\begin{figure} 
\centerline{\includegraphics[scale = 0.6]{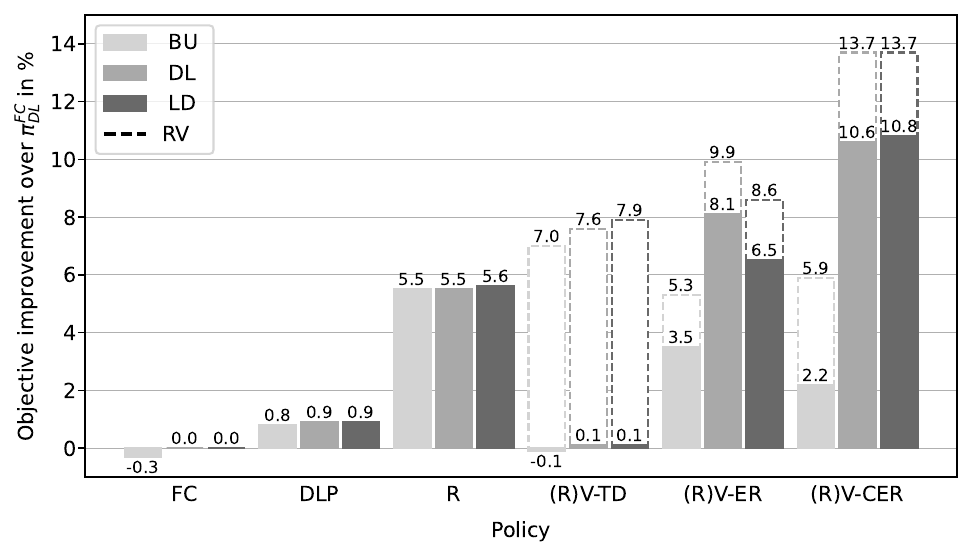}} 
\caption{Objective improvement over myopic benchmark. \label{figure_performance}} 
\caption*{\footnotesize \textit{Note.} $\pi^\text{FC}_\text{LD}$ yields a slight objective decrease of -0.04\% compared to $\pi^\text{FC}_\text{DL}$. Consequently, we declare $\pi^\text{FC}_\text{DL}$ as the best-performing policy with FC. Due to rounding to one decimal place, this is not immediately evident in the figure. For the last three groups of columns, solid bars represent the policies without rollout (V), and bars with dashed outlines depict the variant with rollout (RV).} 
\end{figure}

To evaluate the performance of the proposed policies, we report the average objective improvement across all settings compared to $\pi^\text{FC}_\text{DL}$ in Figure \ref{figure_performance}. Notably, all demand control approaches outperform the myopic benchmark FC, except for $\pi^\text{V-TD}_\text{BU}$. In the subsequent sections, we delve into our main findings.

\subsubsection{Horizon versus Detail.} In line with \cite{soeffker_etal_2022}, we observe a trade-off between the time horizon and the level of detail with which information and dynamics are represented in the demand control approaches. While the DLP covers a relatively long horizon of multiple days, it largely ignores the effects of stochastic information realizing over time. Crucially, it neglects the impact of the feasibility check and is prone to underestimate a request's opportunity cost. By contrast, R depicts dynamics on a very granular level, albeit only for a very short horizon due to the high computational burden. In comparison, the VFA in (R)V-TD, (R)V-ER, and (R)V-CER (implicitly) reflects a much longer horizon but loses some information as a result of the aggregation into features. We can mitigate this by combining the latter two concepts: Figure \ref{figure_performance} shows that all RV policies outperform their respective V counterparts. 

\subsubsection{VFA Variants.} Regarding the tested VFA variants ((R)V-TD, (R)V-ER, (R)V-CER), we find that if we solely rely on TD updates (V-TD), the policies struggle to outperform the benchmark. Interestingly, the combination with a rollout (RV-TD) nevertheless leads to better results than R. Apparently, there is a benefit to combining a rollout with VFA even if the latter is not a very good approximation on its own. 

For capacity-window-based allocation (DL and LD), the approaches with ER ((R)V-ER and (R)V-CER) lead to substantial performance improvements. As an explanation, recall that the VFA features are calculated based on sample realizations of pickups. This introduces randomness into the mapping of post-decision states to features, i.e., a specific state is not always represented by the exact same feature vector. ER updates help to alleviate any destabilizing effects that might arise from our feature design. Enforcing structure in the value function by incorporating domain knowledge through constraints ((R)V-CER) leads to further improvements, making RV-CER the best demand control approach (when combined with DL or LD). 

Remarkably, these observations do not hold for BU. To get to the bottom of this, we highlight two key differences between BU and DL/LD: Firstly, DL and LD both involve lexicographic optimization, which reduces the likelihood of a nonunique tentative allocation plan per scenario. By contrast, BU only focuses on the next allocation decision such that there might regularly exist multiple plans per scenario. This adds a second source of `randomness' to the state-feature-mapping and might destabilize the training to such an extent that ER cannot compensate it. Secondly, DL and LD aim at optimizing capacity windows, which is also the core idea behind the VFA features. As a result, DL and LD are more well-aligned with the VFA.

\textbf{Insight 3:} \textit{The solution framework must take into account interdependencies between decision types}. This holds in both directions: Allocation must match demand control (VFA performs better when combined with DL or LD). Vice versa, demand control must match allocation. To illustrate this, we conduct a supplementary experiment in Appendix \ref{mismatch} where we use a different objective for the VFA features than for allocation. The results show that the performance can deteriorate drastically because of this mismatch, i.e., the VFA cannot handle strongly differing objectives for feature calculation and allocation even if trained on it.

\subsubsection{Demand Control versus Allocation.} Apart from the policies involving ER, the results support our hypothesis that demand control is the stronger lever for improvement. For (R)V-ER and (R)V-CER, the difference in performance for the allocation approaches can be explained with Insight 3.

\subsection{Managerial Analysis} \label{cs_managerial}
Having identified $\pi^\text{RV-CER}_\text{LD}$ as the best policy in terms of overall performance, we next take on a managerial perspective. To keep our exposition concise, we only report the results for $\pi^\text{RV-CER}_\text{LD}$ and our benchmark $\pi^\text{FC}_\text{DL}$.

\subsubsection{Setting.} 
\begin{figure} 
\parbox{.50\linewidth}{{\centerline{\includegraphics[scale = 0.55]{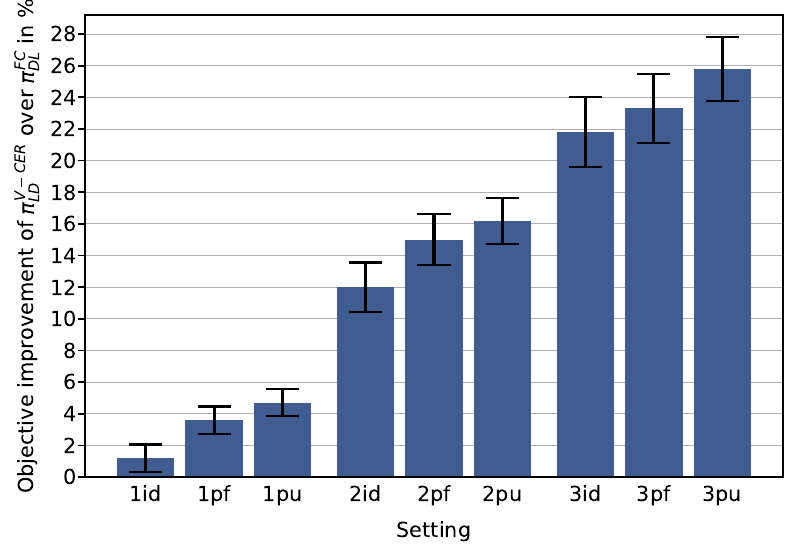}}} 
\caption{Objective improvement per setting. \label{figure_performance_sett}} 
\caption*{\footnotesize \textit{Note.} With 99\% confidence intervals.}} 
\parbox{.50\linewidth}{\vspace{1.1\baselineskip}{\centerline{\includegraphics[scale = 0.25]{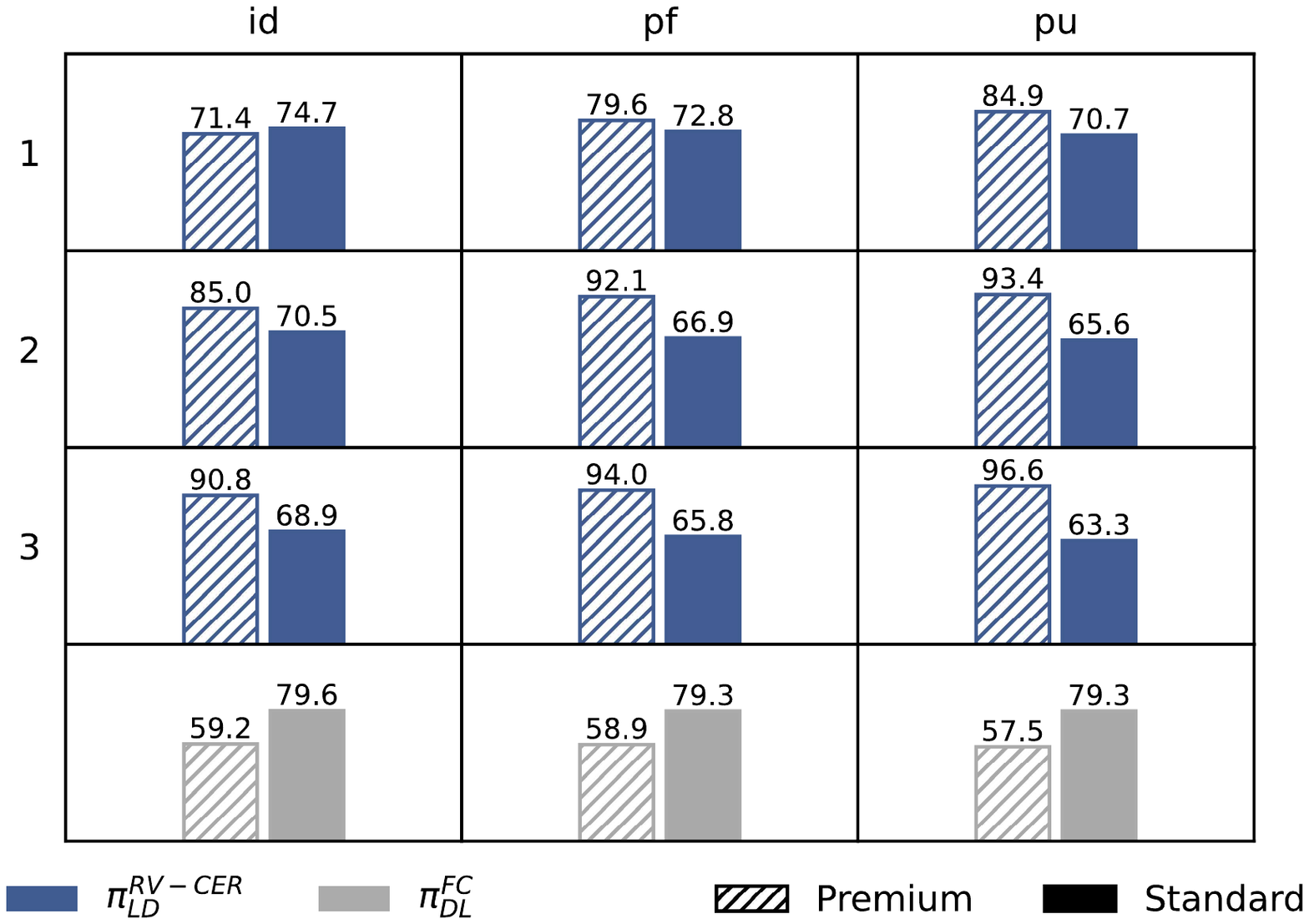}}} 
\vspace{0.45\baselineskip}
\caption{Acceptance rate per customer type. \label{figure_acc_c}} 
\caption*{\footnotesize \textit{Note.} Acceptance rate per customer type in \% for all settings. By design, $\pi^\text{FC}_\text{DL}$ leads to identical results for $m_1\in\{1,2,3\}$.}} 
\end{figure}

Figure \ref{figure_performance_sett} depicts the objective improvement of $\pi^\text{RV-CER}_\text{LD}$ per setting along with the 99\% confidence intervals. In all settings, the improvement is statistically significant at the 1\% significance level, but its extent clearly varies. If the customer types are identical (1id), there still exists a benefit to steering demand, but it is relatively small. In this setting, individual requests merely differ in the parcel size and lead time, such that the displacement effects are comparatively low. The more pronounced the difference in how `valuable' customer types are, the more worthwhile demand management becomes. This difference can be explicit because of the priority weights or implicit due to the pickup behavior (or both).

\textbf{Insight 4:} \textit{The potential gains of demand management increase with the level of heterogeneity between the customer types resulting from the prioritization and/or pickup behavior.}

\subsubsection{Customer Type.} To shed light on the policies' demand control behavior, we report the acceptance rate per customer type and setting in Figure \ref{figure_acc_c}. By design, the control behavior of $\pi^\text{FC}_\text{DL}$ is independent of the customer priority weights, leading to identical results for a given pickup distribution and $m_1\in\{1,2,3\}$. Consequently, we only depict its results once for each distribution. Since $\pi^\text{FC}_\text{DL}$ equals a first-come-first-served policy, its acceptance rate for standard customers is considerably higher than for premium customers, who arrive on short notice. Furthermore, the acceptance rates remain stable across the three pickup distributions because this policy is dominated by standard customers, whose pickup behavior varies only slightly. 

By contrast, $\pi^\text{RV-CER}_\text{LD}$ actively reserves capacity for premium customers, leading to a generally higher acceptance rate for these customers than $\pi^\text{FC}_\text{DL}$. When the customer types are identical (1id), $\pi^\text{RV-CER}_\text{LD}$ produces balanced acceptance rates. If the premium customers become more attractive due to their priority weight or pickup behavior, the balance shifts increasingly in favor of them and the control behavior becomes more aggressive: In setting 3pu, nearly all premium requests are accepted. Figure \ref{figure_performance_sett} demonstrates that $\pi^\text{RV-CER}_\text{LD}$ benefits from exploiting the quicker pickup speed of premium customers in $m_1$pf and $m_1$pu. 

Comparing the two policies, we observe that $\pi^\text{RV-CER}_\text{LD}$ generates a substantial rise in the acceptance rate for premium customers, whereas the decline for standard customers is relatively moderate. In each setting, the boost in the acceptance rate for premium customers overcompensates the decrease for standard customers. Consequently, $\pi^\text{RV-CER}_\text{LD}$ serves between 1.1\% and 4.7\% more customers than $\pi^\text{FC}_\text{DL}$. 

\textbf{Insight 5:} \textit{$\pi^\text{RV-CER}_\text{LD}$ achieves its performance gains mainly by reserving capacity for premium customers. The extent of this control behavior depends on the heterogeneity between customer types.}

\begin{figure}
\centerline{\includegraphics[scale = 0.575]{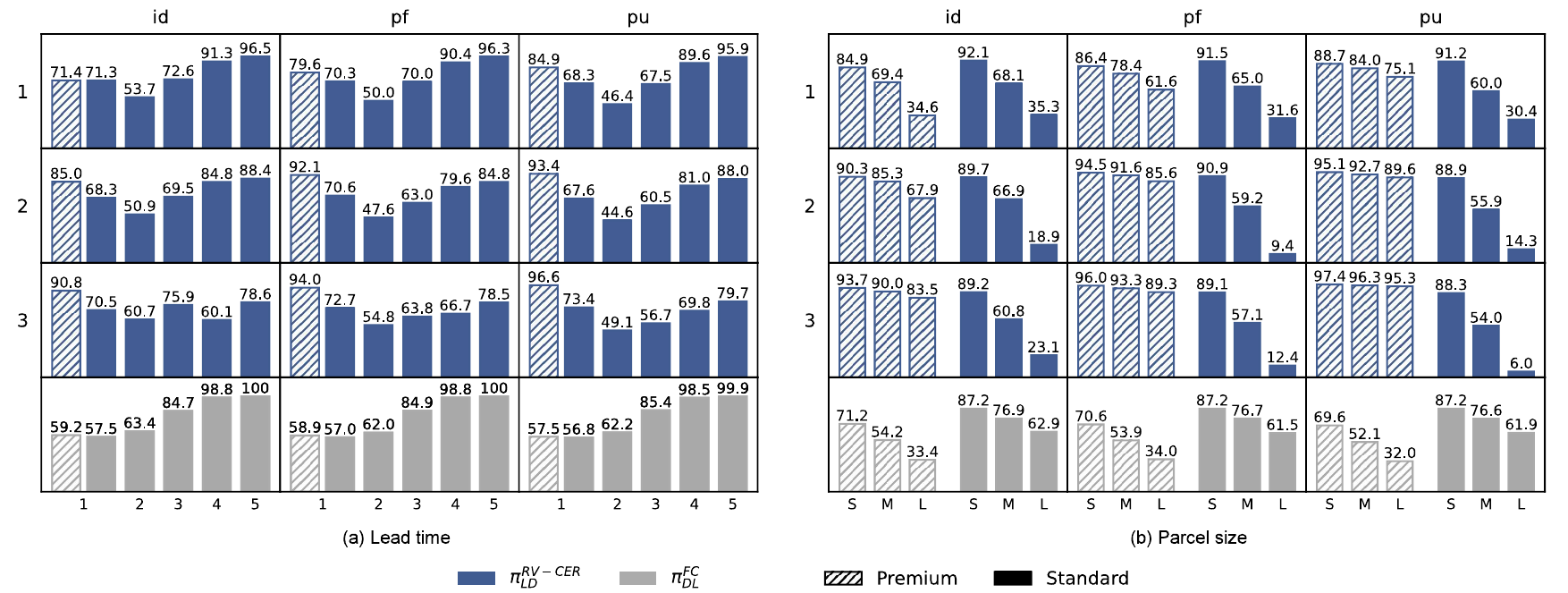}}
\caption{Acceptance rates. \label{figure_accs}}
\caption*{\footnotesize \textit{Note.} Acceptance rate per customer type and lead time (a) or parcel size (b) in \% for all settings. By design, $\pi^\text{FC}_\text{DL}$ leads to identical results for $m_1\in\{1,2,3\}$.}
\end{figure}

\subsubsection{Lead Time.} Digging deeper, we next analyze the acceptance rate per customer type and lead time $e$ in Figure \ref{figure_accs}(a). The first-come-first-served nature of $\pi^\text{FC}_\text{DL}$ becomes evident with acceptance rates of about 100\% for longer lead times that sharply decline for $e\in\{1,2,3\}$. By contrast, $\pi^\text{RV-CER}_\text{LD}$ reduces the acceptance rates for longer lead times and instead favors requests with $e=1$. With increasing heterogeneity between customer types, these effects amplify. Interestingly, $\pi^\text{RV-CER}_\text{LD}$ generates higher acceptance rates for $e=1$ for both customer types. The increase for premium customers is stronger than for standard customers and clearly motivated by their higher value in most settings. Nevertheless, $\pi^\text{RV-CER}_\text{LD}$ also prefers $e=1$ for standard customers. We attribute this observation to two reasons: First, $\pi^\text{RV-CER}_\text{LD}$ aims to maintain flexibility by not committing too early to standard customers and instead reserves capacity for premium customers. However, reserving capacity for premium customers too aggressively means foregoing rewards from accepting standard customers. Consequently, the policy `fills up' remaining capacity with standard customers once it gets more information on the arrival of premium customers. Second, $\pi^\text{RV-CER}_\text{LD}$ strives to keep the amount of time a parcel spends in the system relatively low. Orders with long lead times remain in the system for a longer time and can impact the feasibility check of many subsequent requests.

\subsubsection{Parcel Size.} \label{parcel_size} In Figure \ref{figure_accs}(b), we investigate the acceptance rate per customer type and parcel size. For both policies, the acceptance rate drops with increasing parcel size. This is to be expected given that larger parcels fit less compartments. Matching our findings from Figures \ref{figure_acc_c} and \ref{figure_accs}(a), $\pi^\text{FC}_\text{DL}$ produces lower acceptance rates for premium than for standard customers across all parcel sizes. By comparison, $\pi^\text{RV-CER}_\text{LD}$ drastically raises the acceptance rates for premium customers, especially for medium- and large-sized parcels, with the intensity again depending on how attractive premium customers are compared to standard customers. Clearly, the policy is able to trade off the higher opportunity cost of larger parcels (Insight 1) with the immediate reward of the customer type and hence distinguishes between standard and premium customers. Remarkably, the acceptance rate for small parcels of standard customers under $\pi^\text{RV-CER}_\text{LD}$ tends to be slightly higher than with $\pi^\text{FC}_\text{DL}$. This shows that $\pi^\text{RV-CER}_\text{LD}$ maintains flexibility not only through controlling lead times, but also by favoring small parcels from standard customers as they offer more possibilities for allocation. Overall, Figure \ref{figure_accs}(b) illustrates that standard customers with large parcels mainly `pay the price' for the improvement in the objective in the form of a substantially reduced acceptance rate.

\textbf{Insight 6:} \textit{The control behavior of $\pi^\text{RV-CER}_\text{LD}$ is characterized by preserving flexibility. This manifests itself in high acceptance rates for short lead times and small parcels of standard customers. On the flip side, standard customers with large parcels experience a severely curtailed acceptance rate compared to $\pi^\text{FC}_\text{DL}$.}

\subsubsection{System State.} In Appendix \ref{occupancy+orders}, we track how the system evolves over time in terms of locker occupancy and pending orders. Due to its tendency toward short lead times and small parcels, $\pi^\text{RV-CER}_\text{LD}$ has fewer and less large orders in the system, whereas the number of occupied compartments is similar to $\pi^\text{FC}_\text{DL}$.

\subsubsection{Prioritization.} We conclude this section with a short discussion of the prioritization scheme and its impact. Our results confirm the hypothesis by \cite{sethuraman_etal} that differences in pickup behavior already lead to an implicit prioritization even if all customer types are associated with the same priority weight. To incentivize customers toward faster pickups and increase the throughput of the system, companies could communicate this fact openly to customers. Building on top of this implicit prioritization, the priority weights allow the decision maker to explicitly include further managerial considerations such as customer lifetime value by defining desired acceptance rates and then calibrating the prioritization accordingly through simulation. Given that the prioritization can gravely impact the acceptance rate of certain subgroups, e.g., standard customers with large parcels, it should be designed and monitored with care.

\textbf{Insight 7:} \textit{Differences in pickup speed are reflected in the control behavior of $\pi^\text{RV-CER}_\text{LD}$ and lead to an implicit prioritization. Priority weights are an instrument to incorporate further managerial considerations.}

\section{Conclusion and Outlook} \label{conclusion}
This paper introduces a novel problem to dynamically manage the demand for parcel lockers. From an application perspective, its distinguishing features are uncertain resource usage durations as well as an `upgrading' mechanism due to different compartment sizes. We develop a solution framework which handles the interdependencies between two decision types arising at different points in time. 

As a general methodological contribution, we augment the value function approximation (VFA) training process through a modified version of experience replay (ER). In the style of an ablation study, we quantify the performance contribution of our algorithmic enhancements. While pure temporal difference updates lead to a performance improvement of 0.1\% (7.9\% when combined with a rollout), incorporating ER generates a substantial improvement of 8.1\% (9.9\%) compared to the myopic benchmark. By embedding domain knowledge, we can further increase this to 10.8\% (13.7\%). Our work illustrates that it is worthwhile to enforce structural properties in the value function. 

From a managerial perspective, our results show that the performance gains of our policy increase with the heterogeneity of customer types. Across different settings, we report statistically significant improvements in the objective of up to 26\%. These mainly result from reserving capacity for high-priority customers and preserving flexibility by favoring small parcels and short lead times. Differences in pickup behavior lead to an implicit prioritization that can be refined through explicit priority weights. However, we also show that certain requests (large parcels of low-priority customers) might get rejected disproportionately often.

The conceptual focus of our paper is supported by extensive appendices that provide methodological foundations and implementation details. This includes novel formulations for the decision spaces that enable an intuitive proof of monotonicity in opportunity cost. Building on these, we develop integer programs for the cost function approximation (CFA) decision space and VFA features and devise CFA parameterizations along with corresponding proofs. Our supplementary experiments show that the alignment of demand control and allocation is crucial to performance since a mismatch can lead to decreases of up to 47.5\% compared to the myopic benchmark.

Concluding our work, we identify several promising areas for future research:
\begin{itemize}
    \item \textit{Overbooking.} In our paper, the feasibility check strongly errs on the side of caution. To trade off the two types of classification errors, incorporating an overbooking mechanism is a compelling future extension.
    \item \textit{Product design.} We expect a rise in publications on more sophisticated product designs, such as customer-type-dependent maximum storage times or differentiated prices across parcel sizes.
    \item \textit{Business models.} Many questions pertaining to specific business models remain unexplored. For example, a logistics service provider should consider demand management and vehicle routing integratively.
\end{itemize}

\newpage
\begin{appendices}

\section{Notation} \label{overview_notation} 

\renewcommand{\arraystretch}{1.3}
{\small
\begin{longtable}{p{6.5cm} p{9.2cm}}
$\delta\in\mathcal{D}=\{1,\dots,D\}$         &   Compartment size (indexed in ascending order)\\
$d\in\mathcal{D}$                            &   Parcel size \\
$Q_\delta$                                   &   Number of compartments per size $\delta$ \\
$\tau$                                       &   Day \\
$t\in\mathcal{T}=\{1,\dots,T\}$              &   Point in time \\
$t\in\mathcal{T}^+=\mathcal{T}\cup\{T+1\}$   &   Point in time including allocation decision in $T+1$ \\
$\mathfrak{F}^a$                             &   Request arrival process \\
$c\in\mathcal{C}=\{1,\dots,C\}$              &   Customer type \\
$m_c\in\mathcal{M}$                          &   Priority weight of customer type $c$ \\
$e\in\mathcal{E}=\{1,\dots,E\}$              &   Lead time \\
$\psi=(b,q)$                                 &   Pickup time \\
$b\in\mathcal{B}=\{1,\dots,B\}$              &   Number of days after allocation of pickup \\
$B$                                          &   Maximum storage time \\
$q\in\mathcal{T}$                            &   Point in time of pickup \\
$\mathfrak{F}^p$                             &   Pickup probability distribution \\
$k$                                          &   Decision epoch \\
$S_k=(\tau_k,t_k,L_k,O_k,r_k)\in\mathcal{S}$ &   Pre-decision state in decision epoch $k$ \\
$\tau_k$                                     &   Day in decision epoch $k$ \\
$t_k$                                        &   Point in time in decision epoch $k$ \\
$h\in\mathcal{H}=\{1,\dots,B-1\}$            &   Dwell time \\
$L_k=(l^k_{\delta c h})_{\delta\in\mathcal{D},c\in\mathcal{C},h\in\mathcal{H}}$ & Locker occupancy in pre-decision state in decision epoch $k$ \\
$f\in\mathcal{F}=\{1,\dots,F\}$              &   Remaining fulfillment time \\
$O_k=(o^k_{d c f})_{d\in\mathcal{D},c\in\mathcal{C},f\in\mathcal{F}}$ & Pending orders in pre-decision state in decision epoch $k$ \\
$r\in\mathcal{R}=\{1,\dots,C\cdot D\cdot E\}$ &  Request type with customer type $c_r$, parcel size $d_r$, and lead time $e_r$ \\
$r_k$                                        &   Request type in decision epoch $k$ \\
$S_0$                                        &   Initial state \\                              
$X_k\in\mathcal{X}(S_k)$                        &   Decision in decision epoch $k$ \\
$g_k\in\mathcal{G}(S_k)\subseteq\{0,1\}$     &   Demand control decision in decision epoch $k$ \\
$A_k=(a^k_{d \delta c})_{d\in\mathcal{D},\delta\in\{d,\dots,D\},c\in\mathcal{C}}\in\mathcal{A}(S_k)$ & Allocation decision in decision epoch $k$ \\
$S_k^x=(\tau_k,t_k,L_k^x,O_k^x)\in\mathcal{S}^x$ &   Post-decision state in decision epoch $k$ \\
$L_k^x$                                      &   Locker occupancy in post-decision state in decision epoch $k$ \\
$O_k^x$                                      &   Pending orders in post-decision state in decision epoch $k$ \\
$S^{Mx}(S_k,X_k)$                            &   Transition function for transition from pre-decision state $S_k$ with decision $X_k$ to post-decision state $S_k^x$ \\
$W_{k+1}=(\tau_{k+1},t_{k+1},r_{k+1},P_{k+1}) \in \mathcal{W}(S_k^x$) & Exogenous information \\
$S^{MW}(S_k^x,W_{k+1})$                      &   Transition function for transition from post-decision state $S_k^x$ with exogenous information $W_{k+1}$ to pre-decision state $S_{k+1}$ \\
$\mathbb{P}(W_{k+1}|S_k^x)$                  &   Probability of exogenous information $W_{k+1}$ in post-decision-state $S_k^x$ \\
$P_{k+1} = (p_{\delta ch}^{k+1})_{\delta\in\mathcal{D},c\in\mathcal{C},h\in\mathcal{H}}$ & Pickups during transition from post-decision state $S_k^x$ to pre-decision state $S_{k+1}$ \\
$R(S_k,X_k)$                                 &   Reward function \\
$\pi\in\Pi$                                  &   Policy \\
$X^\pi_k(S_k)$                               &   Decision in pre-decision state $S_k$ when following policy $\pi$ \\
$\pi^*$                                      &   Optimal policy \\
$\bar{R}^*$                                  &   Reward rate of the optimal policy \\
$V(S)$                                       &   Value function in state $S\in\mathcal{S}\cup\mathcal{S}^x$ \\
$\Delta V(S_k^x)$                            &   Opportunity cost in decision epoch $k$ \\
$\bm{\phi}(S_k^x)$                              &   Feature vector for post-decision state $S_k^x$ \\
$\bm{\theta}$                                   &   Vector of trained weights for value function approximation \\
$\lambda$                                    &   Length of capacity window \\
$w_{\delta \lambda}$                         &   Number of capacity windows in tentative allocation plan in a compartment of size $\delta$ with length $\lambda$ \\
$v_{\delta\lambda}$                          &   Objective function coefficients for cost function approximation \\
$\hat{V}(S|\bm{\theta})$                        &   Approximated value function \\
$u=1,\dots,U$                                &   Scenarios for features \\
$\hat{\mathcal{F}}=\{1,\dots,F+B-1\}$        &   Planning horizon for tentative allocation plans of features \\
$\hat{w}_{\delta\lambda}^u$                  &   Number of capacity windows per compartment size $\delta$ and length $\lambda$ in scenario $u$ \\
$\hat{w}_{\delta\lambda}$                    &   Capacity window features for value function approximation \\
$\tau^\text{max}$                            &   Number of simulated days for training \\
$n$                                          &   Simulated decision epoch during training \\
$\varepsilon_n$                              &   Exploration scheme \\
$\mathbb{M}$                                 &   Replay memory \\
$\kappa$                                     &   Size of the replay memory \\
$\chi$                                       &   Size of the samples for experience replay updates \\
$\eta$                                       &   Start of experience replay updates \\
$\zeta$                                      &   Frequency of experience replay updates \\
$S_0^x$                                      &   Initial post-decision state \\
$\bar{\mathbb{M}}$                           &   Simple random sample drawn from $\mathbb{M}$ \\
$y_{\delta cf}$                              &   Number of compartments of size $\delta$ with tentatively allocated parcel of customer type $c$ in allocation decision $f$ \\
$\Tilde{O}_k$                                &   Modified pending orders in decision epoch $k$ \\
$V^\pi(S)$                                   &   Value function in state $S$ when following policy $\pi$ \\
$s_{\delta f}$                               &   Number of empty compartments of size $\delta$ in allocation decision $f$ \\
$w_{\delta\lambda f}$                        &   Number of capacity windows in compartment size $\delta$ and with length $\lambda$ that start with allocation decision $f$ \\
$\beta$                                      &   Sampled values for number of days after allocation of pickup \\
$\hat{o}^{ukx}_{\delta cf\beta}$             &   Pending orders with sampled pickup $\beta$ in scenario $u$ \\
$\hat{l}^{ukx}_{\delta ch\beta}$             &   Locker occupancy with sampled pickup $\beta$ in scenario $u$ \\
$\hat{s}_{\delta f}^u$                       &   Number of unoccupied compartments of size $\delta$ in allocation decision $f$ in scenario $u$ \\
$\hat{w}_{\delta\lambda f}^u$                &   Number of capacity windows in compartment size $\delta$ and with length $\lambda$ that start with allocation decision $f$ in scenario $u$ \\
$\hat{y}_{\delta cf\beta}^u$                 &   Number of compartments of size $\delta$ with tentatively allocated parcel of customer type $c$ in allocation decision $f$ with sampled pickup $\beta$ in scenario $u$ \\
$\alpha^\theta$                              &   Stepsize for temporal difference learning updates \\
$\alpha^{\bar{R}}, \alpha_n^{\bar{R}}$       &   Stepsizes for reward rate estimate \\
$\bar{R}$                                    &   Reward rate estimate \\
$\gamma$                                     &   Ridge regression regularization parameter \\
$\rho$                                       &   Auxiliary variable for reward rate estimate \\
$\Delta$                                     &   Temporal difference error \\
$(\bm{\phi}^0_n,\bm{\phi}^\text{reject}_n,\bm{\phi}^\text{accept}_n,R^\text{accept}_n)$ & Experience in simulated decision epoch $n$ \\
$\bm{\phi}^0_n$                                 &   Feature vector in simulated decision epoch $n-1$ \\
$\bm{\phi}^\text{reject}_n$                     &   Feature vector after rejecting in simulated decision epoch $n$ \\
$\bm{\phi}^\text{accept}_n$                     &   Feature vector after accepting in simulated decision epoch $n$ \\
$R^\text{accept}_n$                         &   Reward of accepting in simulated decision epoch $n$ \\
$\nu_i$                                      &   Prediction target for experience replay \\
$\bar{\bm{\phi}}^0_i$                           &   Standardized features \\
$\bm{\theta}^{{ER^*}}$                          &   Standardized feature weight vector \\
$\Omega$                                     &   Number of sample paths for truncated online rollout\\
$\xi$                                        &   Horizon of truncated online rollout \\
$\varphi^\text{DLP}$                         &   Horizon of deterministic linear program \\
$\vartheta,\varphi\in\{1,\dots,\varphi^\text{DLP}\}$ & Day in the deterministic linear program \\
$\hat{o}_{dc\varphi}$                        &   Expected number of future requests with parcel size $d$, customer type $c$, delivery date $\varphi$ \\
$\bar{p}_{ch\varphi}$                        &   Probability that a parcel of customer type $c$ with dwell time $h$ on day $\tau_k$ is still in the locker on day $\varphi$ \\
$\hat{p}_{c\vartheta\varphi}$                &   Probability that a parcel of customer type $c$ allocated on day $\vartheta$ is still in the locker on day $\varphi$
\end{longtable}}

\section{Decision spaces} \label{MIP_feasibility_check}
In the following, we state the constraints to determine the decision spaces for demand control and allocation. This requires additional notation: Let decision variable $y_{\delta cf}$ denote the number of compartments of size $\delta$ to which we tentatively allocate a parcel belonging to a customer of type $c$ in the $f$\textsuperscript{th} allocation decision from now (with $f=1$ referring to the allocation decision on the current day $\tau_k$). 

To specify the demand control decision space, we must perform the feasibility check. We model the acceptance of the current request $r_k$ by modifying $O_k$ to $\Tilde{O}_k$ with $\Tilde{o}^k_{d c f}=o^k_{d c f}+\mathbf{1}(d=d_{r_k}, c=c_{r_k},f=f_{r_k})$ and $\mathbf{1}(\cdot)$ symbolizing the indicator function. The feasibility check then reduces to determining whether a solution satisfying the following constraints exists:
\begin{align}
    & \sum_{c\in\mathcal{C}} \sum_{j=1}^f y_{\delta c j} + \sum_{c\in\mathcal{C}} \sum_{h=1}^{B-f} l_{\delta c h}^k \leq Q_\delta && \delta \in \mathcal{D},\ f = 1,\dots,\min\{F,B-1\} \label{cap1}\\
    & \sum_{c\in\mathcal{C}}\sum_{j=0}^{B-1}y_{\delta c,f-j} \leq Q_\delta && \delta \in \mathcal{D},\ B \leq f \leq F \label{cap2}\\
    & \sum_{j=1}^\delta y_{jcf} \leq \sum_{d=1}^\delta \Tilde{o}_{dcf}^k && \delta \in \mathcal{D}\setminus\{D\},\ c\in\mathcal{C},\ f\in\mathcal{F} \label{order_mix} \\
    & \sum_{\delta\in\mathcal{D}} y_{\delta cf} = \sum_{d\in\mathcal{D}} \Tilde{o}_{dcf}^k && c\in\mathcal{C},\ f\in\mathcal{F} \label{all} \\
    & y_{\delta cf} \in \mathbb{N}_0 && \delta \in \mathcal{D},\ c\in\mathcal{C},\ f\in\mathcal{F} \label{int}
\end{align}

Constraints (\ref{cap1}) and (\ref{cap2}) ensure that the number of compartments is not exceeded, taking into account the current occupancy of the locker, the tentative allocation decisions over time as well as the maximum storage time. Constraints~(\ref{order_mix}) guarantee that orders are assigned to compatible compartments. More specifically, they ensure that the number of occupied compartments up to a specific size $\delta$ does not exceed the number of eligible orders, i.e., orders with a parcel of size $d\leq\delta$, while simultaneously allowing for the upgrading of orders to larger compartment sizes. Constraints (\ref{all}) state that all orders in $\Tilde{O}_k$ must be allocated.

In other words, the feasibility check can be interpreted as a constraint satisfaction problem defined by (\ref{cap1})-(\ref{int}). If a solution exists, the request is classified as feasible and the decision maker must choose between rejection or acceptance ($\mathcal{G}(S_k)=\{0,1\}$). Otherwise, the request is infeasible and must be rejected ($\mathcal{G}(S_k)=\{0\}$).

For the allocation decision space, we require constraints (\ref{cap1}) to (\ref{int}) with $\Tilde{O}_k=O_k$ along with two additional constraints linking $a^k_{d \delta c}$ to $y_{\delta c1}$ and $O_k$:
\begin{align}
    & y_{\delta c 1} = \sum^\delta_{d=1} a^k_{d\delta c} && \delta\in\mathcal{D},\ c\in\mathcal{C} \label{allocate1}\\
    & o^k_{dc1} = \sum^D_{\delta=d} a^k_{d\delta c} && d\in\mathcal{D},\ c\in\mathcal{C} \label{allocate2}\\
    & a_{d\delta c}^k \in \mathbb{N}_0 && d \in \mathcal{D},\ \delta\in\{d,\dots,D\},\ c\in\mathcal{C} \label{int_a}
\end{align}

Constraints (\ref{allocate1}) require that the number of compartments of size $\delta$ that are used in the tentative plan for customers of type $c$ in $f=1$ must match the number of compatible parcels with size $d\leq\delta$ that are allocated to compartments of size $\delta$ in the current allocation decision represented by $a^k_{d\delta c}$. Constraints (\ref{allocate2}) state that all orders of size $d$ and customer type $c$ with remaining fulfillment time $f=1$ must be allocated to compatible compartments with size $\delta\geq d$ in the current allocation decision.

\section{Proof for monotonicity of opportunity cost} \label{monotonicity}
This section is dedicated to the proof that a request's opportunity cost is nondecreasing with increasing parcel size. In preparation, we briefly restate the opportunity cost definition from Section \ref{sdp_valuefunc}:
\begin{equation} \label{oc}
    \Delta V(S_k) = V(S^{Mx}(S_k,g_k=0)) - V(S^{Mx}(S_k,g_k=1))
\end{equation}

For the proof, we consider two arbitrary demand control pre-decision states $\dot{S}_k$ and $\bar{S}_k$ in a specific decision epoch $k$ that only differ in their current request's parcel size. All other components of the states are identical, i.e., $\dot{\tau}_k=\bar{\tau}_k,\dot{t}_k=\bar{t}_k,\dot{L}_k=\bar{L}_k,\dot{O}_k=\bar{O}_k$. For the sake of brevity, we denote the request's parcel size by $\dot{d}$ and $\bar{d}$ with $\dot{d}<\bar{d}$ for the respective states $\dot{S}_k$ and $\bar{S}_k$. The two other request attributes, i.e., the customer type and lead time, do not differ between the two states and are represented by $\ddot{c}$ and $\ddot{f}$. Drawing on the definition given by (\ref{oc}), the monotonicity property translates to:
\begin{gather}
    \Delta V(\bar{S}_k) \geq \Delta V(\dot{S}_k) \label{monotonicity_property1} \\
    V(S^{Mx}(\bar{S}_k,g_k=0)) - V(S^{Mx}(\bar{S}_k,g_k=1)) \geq V(S^{Mx}(\dot{S}_k,g_k=0)) - V(S^{Mx}(\dot{S}_k,g_k=1)) \label{monotonicity_property2}
\end{gather}

Note that the values in case of rejection ($g_k=0$) are equal as we reach exactly the same post-decision state in both cases. Consequently, to prove (\ref{monotonicity_property2}) and thus (\ref{monotonicity_property1}), it suffices to show that:
\begin{equation} \label{m_values}
    V(S^{Mx}(\dot{S}_k,g_k=1))\geq V(S^{Mx}(\bar{S}_k,g_k=1))
\end{equation}

To prove (\ref{m_values}), we first observe that the post-decision states $\dot{S}_k^x$ and $\bar{S}_k^x$ which result from accepting the request in $\dot{S}_k$ and $\bar{S}_k$ respectively only differ with respect to the pending orders. In light of this, we can directly construct $\bar{S}^x_k$ from $\dot{S}^x_k$ by adjusting the orders according to $\bar{o}^{kx}_{\dot{d}\ddot{c}\ddot{f}}=\dot{o}^{kx}_{\dot{d}\ddot{c}\ddot{f}} - 1$ and $\bar{o}^{kx}_{\bar{d}\ddot{c}\ddot{f}}=\dot{o}^{kx}_{\bar{d}\ddot{c}\ddot{f}} + 1$ with $\bar{o}^{kx}_{dcf} = \dot{o}^{kx}_{dcf}$ in all other cases. Note that the exogenous information $W_{k+1}$ and its probability distribution are not affected. Consequently, the transition from the post-decision states $\dot{S}_k^x$ and $\bar{S}_k^x$ to the subsequent pre-decision states $\dot{S}_{k+1}$ and $\bar{S}_{k+1}$ is unaltered and, for each realization of exogenous information, the two pre-decision states again only differ with regard to the pending orders.

Importantly, this difference in pending orders affects the decision spaces $\mathcal{X}(\dot{S}_{k+1})$ and $\mathcal{X}(\bar{S}_{k+1})$ in $k+1$. As a key result, we show that every decision in $\mathcal{X}(\bar{S}_{k+1})$ is also a feasible decision for $\mathcal{X}(\dot{S}_{k+1})$, but not vice versa:
\begin{equation} \label{subset_ds}
    \mathcal{X}(\bar{S}_{k+1})\subset \mathcal{X}(\dot{S}_{k+1})
\end{equation}

The constraints that shape the decision spaces are formalized in Appendix \ref{MIP_feasibility_check}. For both demand control and allocation, the impact of the different pending orders in the two pre-decision states is limited to constraints (\ref{order_mix}) and (\ref{all}):
\begin{itemize}
    \item Comparing $\mathcal{X}(\dot{S}_{k+1})$ and $\mathcal{X}(\bar{S}_{k+1})$, constraint (\ref{order_mix}) only differs for $c=\ddot{c},\ f=\ddot{f},\ \delta \in \{\dot{d},\dots,\bar{d}-1\}$. We can construct the adapted constraint for $\mathcal{X}(\bar{S}_{k+1})$ from the corresponding one for $\mathcal{X}(\dot{S}_{k+1})$:
    \begin{align}
        \sum_{j=1}^\delta y_{j\ddot{c}\ddot{f}} \leq \sum_{j=1}^\delta \dot{o}_{j\ddot{c}\ddot{f}}^k - 1 && \delta \in \{\dot{d},\dots,\bar{d}-1\} \label{order_mix_altered}
    \end{align}
    For $\delta\in\{\bar{\delta},\dots,D-1\}$, the reduction in $\bar{o}^{kx}_{\dot{d}\ddot{c}\ddot{f}}$ and the increase in $\bar{o}^{kx}_{\bar{d}\ddot{c}\ddot{f}}$ for $\bar{S}_{k+1}$ compared to $\dot{S}_{k+1}$ cancel each other out in the summation, yielding the exact same constraint as for $\mathcal{X}(\dot{S}_{k+1})$.
    \item Regarding (\ref{all}), we observe that the total number of orders for $\dot{S}_{k+1}$ and $\bar{S}_{k+1}$ is identical. 
\end{itemize}
\newpage
As a result from these two observations, we conclude that the decision spaces only differ with regard to a reduced right-hand side in constraint (\ref{order_mix}) for $\mathcal{X}(\bar{S}_{k+1})$ as illustrated in (\ref{order_mix_altered}). Consequently, every feasible decision out of $\mathcal{X}(\bar{S}_{k+1})$ is automatically also a feasible decision for the less restricted decision space $\mathcal{X}(\dot{S}_{k+1})$, but not vice versa.

Repeatedly applying this argument, we can feasibly replicate the sequence of decisions dictated by the optimal policy $\pi^*$ starting from $\bar{S}_k^x$ and implement it from $\dot{S}_k^x$ onward, yielding a new policy $\bar{\pi}$. By following $\bar{\pi}$ from $\dot{S}_k^x$, we thus behave as if we were applying $\pi^*$ starting from $\bar{S}_k^x$. The expected sequence of rewards is encapsulated in the value function and must be identical in both cases. Letting $V^\pi(S)$ denote the value if we apply policy $\pi$ from state $S$ onward, we obtain $V^{\bar{\pi}}(S^{Mx}(\dot{S}_k,g_k=1))=V^{\pi^*}(S^{Mx}(\bar{S}_k,g_k=1))$. By contrast, the value $V^{\pi^*}(S^{Mx}(\dot{S}_k,g_k=1))$ resulting from applying the optimal policy $\pi^*$ from $\dot{S}^x_k$ onward assumes that we select the optimal, value-maximizing decision in every subsequent decision epoch. As a general property, $V^{\pi^*}(S)\geq V^{\pi}(S)$ holds for all states and policies \citep{mahadevan_1996}. Therefore, $V(S^{Mx}(\dot{S}_k,g_k=1)) = V^{\pi^*}(S^{Mx}(\dot{S}_k,g_k=1))\geq V^{\bar{\pi}}(S^{Mx}(\dot{S}_k,g_k=1)) = V(S^{Mx}(\bar{S}_k,g_k=1))$, which proves (\ref{m_values}) and thereby (\ref{monotonicity_property1}).

\section{Cost function approximation: Implementation details} \label{CFA_details}
In this section, we shed light on the details for implementing the cost function approximation (CFA). First, we formalize the CFA decision space. Second, we elaborate on the parameterizations for the CFA objective. Lastly, we demonstrate how the lexicographic optimization can be implemented efficiently by appropriately weighting the two objective functions.

\subsection{CFA decision space} \label{cfa_decision_space}
To model the CFA decision space, we require auxiliary decision variables $s_{\delta f}$ to denote the number of unoccupied compartments of size $\delta$ in allocation decision $f$, $w_{\delta\lambda f}$ as the number of capacity windows in a compartment of size $\delta$ and with length $\lambda$ that start with allocation decision $f$, and $y_{\delta cf}$ as the number of compartments of size $\delta$ that are allocated a parcel belonging to a customer of type $c$ in $f$ (see also Appendix \ref{MIP_feasibility_check}). We model the CFA decision space with the following constraints:
\newpage
\begin{align}
    & \sum_{c\in\mathcal{C}} \sum_{j=1}^f y_{\delta c j} + \sum_{c\in\mathcal{C}} \sum_{h=1}^{B-f} l_{\delta c h}^k + s_{\delta f} = Q_\delta && \delta \in \mathcal{D},\ f = 1,\dots,\min\{F,B-1\} \label{CFA_cap1}\\
    & \sum_{c\in\mathcal{C}}\sum_{j=0}^{B-1}y_{\delta c,f-j} + s_{\delta f} = Q_\delta && \delta \in \mathcal{D},\ B \leq f \leq F \label{CFA_cap2}\\
    & \sum_{j=1}^\delta y_{jcf} \leq \sum_{d=1}^\delta o_{dcf}^k && \delta \in \mathcal{D}\setminus\{D\},\ c\in\mathcal{C},\ f\in\mathcal{F} \label{CFA_order_mix} \\
    & \sum_{\delta\in\mathcal{D}} y_{\delta cf} = \sum_{d\in\mathcal{D}} o_{dcf}^k && c\in\mathcal{C},\ f\in\mathcal{F} \label{CFA_all} \\
    & y_{\delta c 1} = \sum^\delta_{d=1} a^k_{d\delta c} && \delta\in\mathcal{D},\ c\in\mathcal{C} \label{CFA_allocate1}\\
    & o^k_{dc1} = \sum^D_{\delta=d} a^k_{d\delta c} && d\in\mathcal{D},\ c\in\mathcal{C} \label{CFA_allocate2}\\
    & \sum_{j=1}^{f}\sum_{\lambda=f-j+1}^{F-j+1} w_{\delta\lambda j} = s_{\delta f} && \delta\in\mathcal{D},\ f\in\mathcal{F} \label{CFA_s_to_w} \\
    &\sum_{j=1}^{f-1} w_{\delta j, f-j} \leq \sum_{c\in\mathcal{C}} y_{\delta cf} && \delta\in\mathcal{D},\ f\in\{2,\dots,F\} \label{CFA_w_end} \\
    & w_{\delta\lambda} = \sum_{f=1}^{F-\lambda+1} w_{\delta\lambda f} && \delta\in\mathcal{D},\ \lambda\in\mathcal{F} \label{CFA_wf_to_w} \\
    & a_{d\delta c}^k,s_{\delta f},\ w_{\delta \lambda},\ w_{\delta\lambda f},\ y_{\delta cf} \in \mathbb{N}_0 && \delta \in \mathcal{D},\ d\in\{1,\dots,\delta\},\ c\in\mathcal{C},f\in\mathcal{F}, \label{CFA_int}\\
    & && \lambda\in\{1,\dots,F-f+1\} \notag
\end{align}

Constraints (\ref{CFA_cap1})-(\ref{CFA_allocate2}) essentially serve the same purpose as constraints (\ref{cap1})-(\ref{allocate2}) in Appendix \ref{MIP_feasibility_check} and model the basic allocation decision space. Note that constraints (\ref{CFA_cap1}) and (\ref{CFA_cap2}) are slightly modified because we must include $s_{\delta f}$ as an auxiliary variable to capture the unused capacity per compartment size $\delta$ and allocation decision $f$. Constraints (\ref{CFA_s_to_w})-(\ref{CFA_wf_to_w}) are needed to correctly determine the number of capacity windows represented by $w_{\delta f}$. More specifically, constraints (\ref{CFA_s_to_w}) ensure that the number of capacity windows that include $f$ matches the available capacity in $f$. Constraints (\ref{CFA_w_end}) require that a capacity window ending before $F$ has to be followed by an occupied compartment. Otherwise, a capacity window of length seven could, for example, also be cast as seven capacity windows of length one. Constraints (\ref{CFA_wf_to_w}) aggregate $w_{\delta\lambda f}$ to $w_{\delta\lambda}$.
\newpage
\subsection{CFA objectives} \label{cfa_param_w}
We propose two objectives for the CFA:
\begin{enumerate}
    \item With the following, we maximize the number of capacity windows in hierarchical order of the compartment size (the largest size having the highest priority): 
    \begin{equation}
        \max\sum_{\delta\in\mathcal{D}}\sum_{\lambda\in\mathcal{F}} \delta\lambda w_{\delta \lambda}\label{cfa_obj_s} 
    \end{equation}
    Basically, the term $\sum_{\lambda\in\mathcal{F}}\lambda w_{\delta \lambda}$ captures the available capacity per compartment size. The vacant capacity is then weighted by the compartment size index $\delta$ to ensure that we prioritize it accordingly. 
    \item To maximize the number of capacity windows in hierarchical order of the window length (with the longest one having the highest priority), we formulate the following objective:
    \begin{equation}
        \max\sum_{\delta\in\mathcal{D}} \sum_{\lambda\in\mathcal{F}} (2\lambda-1)w_{\delta\lambda} \label{cfa_obj_w}
    \end{equation}
\end{enumerate}

While the first objective is trivial, the second one requires more elaboration. In the following, we prove that setting the coefficients in the objective to $2\lambda-1$ leads to a prioritization of capacity windows in hierarchical order of their window length, i.e., we maximize the number of windows in lexicographic order of their length with the longest possible length $\lambda=F$ having the highest priority. Essentially, the proof is based on showing that splitting up a single capacity window $\bar{W}$ in a compartment of size $\bar{\delta}$ and length $\bar{\lambda}\geq 2$ into an arbitrary combination of capacity windows of shorter length leads to a deterioration of the objective value given by (\ref{cfa_obj_w}).

To formalize the notion of splitting up $\bar{W}$, we let $\bar{w}_{\delta\lambda}$ denote the number of capacity windows in a compartment of size $\delta\in\mathcal{D}$ and length $\lambda\in\mathcal{F}$ that $\bar{W}$ is divided into. For a decomposition of $\bar{W}$ to be permissible, it must adhere to the following three constraints:
\begin{align}
    &\sum_{\delta\in\mathcal{D}}\sum_{\lambda\in\mathcal{F}}\lambda \bar{w}_{\delta\lambda}=\bar{\lambda} && \label{total_length} \\
    & \sum_{\delta\in\mathcal{D}}\sum_{\lambda\in\mathcal{F}}\bar{w}_{\delta\lambda} \geq 2 \label{split_at_least_two} \\
    & \bar{w}_{\delta\lambda}\in\mathbb{N}_0 & \delta\in\mathcal{D},\ \lambda\in\mathcal{F} \label{w_int}
\end{align}

Constraints (\ref{total_length}) state that the sum of the length of the shorter windows must in total equal the length of the original window $\bar{W}$ given by $\bar{\lambda}$. Constraints (\ref{split_at_least_two}) require that a decomposition must consist of at least two windows. Lastly, the number of windows $\bar{w}_{\delta\lambda}$ must be integer.

We now show that any decomposition fulfilling constraints (\ref{total_length})-({\ref{w_int}}) leads to a smaller contribution to the objective function than the original capacity window $\bar{W}$:
\newpage
\begin{gather}
    2\bar{\lambda}-1>\sum_{\delta\in\mathcal{D}}\sum_{\lambda\in\mathcal{F}}(2\lambda-1)\bar{w}_{\delta\lambda} \label{cfa_step_1}\\
    (2\sum_{\delta\in\mathcal{D}}\sum_{\lambda\in\mathcal{F}}\lambda\bar{w}_{\delta\lambda})-1>\sum_{\delta\in\mathcal{D}}\sum_{\lambda\in\mathcal{F}}2\lambda\bar{w}_{\delta\lambda} - \sum_{\delta\in\mathcal{D}}\sum_{\lambda\in\mathcal{F}}\bar{w}_{\delta\lambda} \label{cfa_step_2}\\
    \sum_{\delta\in\mathcal{D}}\sum_{\lambda\in\mathcal{F}}\bar{w}_{\delta\lambda} > 1 \label{cfa_step_3}    
\end{gather}

Inequality (\ref{cfa_step_1}) states our hypothesis. Plugging in constraint (\ref{total_length}) and slightly rearranging the terms, we obtain (\ref{cfa_step_2}) and we end up with (\ref{cfa_step_3}). The latter is always true because of (\ref{split_at_least_two}). This proves that (\ref{cfa_step_1}) holds for all decompositions adhering to (\ref{total_length})-(\ref{w_int}).

Given that the total available capacity in the tentative allocation plan is fixed, (\ref{cfa_obj_w}) means that a longer window will always dominate an arbitrary combination of shorter windows. As a result, the objective prioritizes maximizing the number of windows with maximum length $F$ as a primary objective, followed by maximizing the number of windows with length $F-1$ as a secondary objective, and so forth, thereby inducing the desired allocation scheme. 

\subsection{Lexicographic optimization of (\ref{cfa_obj_s}) and (\ref{cfa_obj_w}) and CFA parameterizations}
In this section, we show how the lexicographic optimization of (\ref{cfa_obj_s}) and (\ref{cfa_obj_w}) can be efficiently implemented by adequately weighting the two objective functions. The argumentation is based on upper bounds of the objective values of (\ref{cfa_obj_s}) and (\ref{cfa_obj_w}), which we state in a first step:
\begin{gather}
    1+\sum_{\delta\in\mathcal{D}}\delta FQ_{\delta} \label{ub_S} \\
    1+(2F-1)\cdot\sum_{\delta\in\mathcal{D}} Q_{\delta} \label{ub_W} 
\end{gather}

In both cases, the highest attainable objective value arises if the tentative allocation plan contains no orders, thereby offering the maximum available capacity. In (\ref{ub_S}), this means that the total number of compartments given by $\sum_{\delta\in\mathcal{D}}Q_\delta$ is unoccupied over the entire horizon $F$. For (\ref{ub_W}), we follow a similar argument and additionally use the fact that longer capacity windows dominate shorter windows by design (Appendix \ref{cfa_param_w}). To obtain bounds strictly larger than the highest attainable objective value, we add a constant of 1.

Given that the objective function coefficients in (\ref{cfa_obj_s}) and (\ref{cfa_obj_w}) as well as the decision variables are integer, the smallest possible change in the objectives is 1 unit. If we weight the secondary objective with the reciprocal of its upper bound, the contribution of the secondary objective in the weighted objective function is guaranteed to be strictly less than 1. As a result, we obtain the following two objectives for maximizing (\ref{cfa_obj_s}) and (\ref{cfa_obj_w}) in lexicographic order:
\newpage
\begin{align}
    & \max\sum_{\delta\in\mathcal{D}}\sum_{\lambda\in\mathcal{F}} \delta\lambda w_{\delta \lambda} + \sum_{\delta\in\mathcal{D}}\sum_{\lambda\in\mathcal{F}} \frac{ (2\lambda-1)}{1+(2F-1)\cdot\sum_{\delta\in\mathcal{D}} Q_{\delta}} w_{\delta \lambda} \label{obj_s_first}\\
    & \max\sum_{\delta\in\mathcal{D}} \sum_{\lambda\in\mathcal{F}} (2\lambda-1)w_{\delta\lambda} + \sum_{\delta\in\mathcal{D}}\sum_{\lambda\in\mathcal{F}}\frac{ \delta \lambda}{1+F\cdot\sum_{\delta\in\mathcal{D}}\delta Q_{\delta}}w_{\delta \lambda} \label{obj_w_first}
\end{align}

Weighted objective (\ref{obj_s_first}) treats (\ref{cfa_obj_s}) as the primary and (\ref{cfa_obj_w}) as the secondary objective, (\ref{obj_w_first}) vice versa. They correspond to the following parameterizations of (\ref{cfa_obj}) respectively:
\begin{align}
    & v_{\delta\lambda} = \delta\lambda + \frac{ (2\lambda-1)}{1+(2F-1)\cdot\sum_{\delta\in\mathcal{D}} Q_{\delta}} \\
    & v_{\delta\lambda} = 2\lambda-1 + \frac{ \delta \lambda}{1+F\cdot\sum_{\delta\in\mathcal{D}}\delta Q_{\delta}}w_{\delta \lambda} 
\end{align}

Note that for a extensive maximum lead time $F$, a huge number of compartments $\sum_{\delta\in\mathcal{D}}Q_\delta$ or numerous compartment sizes $D$, the coefficients of the respective secondary objective in (\ref{obj_s_first}) and (\ref{obj_w_first}) might shrink and potentially give rise to numerical problems. In that case, one would have to resort to sequentially optimizing the individual objectives (\ref{cfa_obj_s}) and (\ref{cfa_obj_w}) while adding adequate constraints to implement the lexicographic optimization. 

\section{Value function approximation: Implementation details} \label{VFA_details}
In this section, we elaborate on the implementation of the VFA. We provide details on the feature calculation, the training procedure and the structure-enforcing constraints during experience replay.

\subsection{Feature calculation} \label{VFA_feature_calculation_details}
To generate scenarios of pickup behavior for the VFA features, it is sufficient to sample $b$, i.e., the number of days after allocation until the pickup occurs, because the tentative plan considers each day on an aggregate level, rendering the specific point in time of the pickup irrelevant. Letting $\beta$ denote the sampled values for $b$, we transform the pending orders $o^{kx}_{dcf}$ and orders in the locker $l^{kx}_{\delta ch}$ into $\hat{o}^{ukx}_{dcf\beta}$ and $\hat{l}^{ukx}_{\delta ch\beta}$.

The integer program to calculate the VFA features is a slightly extended version of the CFA decision space presented in Appendix \ref{cfa_decision_space}. The adaptions basically serve to accommodate the sampled pickup behavior. To determine the number of capacity windows $\hat{w}_{\delta\lambda}^u$ with length $\lambda$ and compartment size $\delta$ in the tentative allocation plan with sampled pickups in scenario $u$, we introduce auxiliary decision variables: $\hat{s}_{\delta f}^u$ denotes the number of unoccupied compartments of size $\delta$ in allocation decision $f$ and scenario $u$, $\hat{w}_{\delta\lambda f}^u$ represents the number of capacity windows in a compartment of size $\delta$ and with length $\lambda$ that start with allocation decision $f$ in scenario $u$, and $\hat{y}_{\delta cf\beta}^u$ equals the number of compartments of size $\delta$ to which we allocate a parcel belonging to a customer of type $c$ in $f$ with the pickup occurring $\beta$ days after the allocation in scenario $u$.
\begin{align}
    & \sum_{c\in\mathcal{C}} \sum_{j=1}^f \sum_{\beta=f-j+1}^B \hat{y}_{\delta c j \beta}^u + \sum_{c\in\mathcal{C}} \sum_{h=1}^{B-f} \sum_{\beta=f+h}^{B}\hat{l}_{\delta c h \beta}^{ukx} + \hat{s}_{\delta f}^u = Q_\delta && \delta \in \mathcal{D},\ f = 1,\dots,\min\{F,B-1\} \label{VFA_cap1}\\
    & \sum_{c\in\mathcal{C}}\sum_{j=0}^{B-1}\sum_{\beta=j+1}^B \hat{y}_{\delta c,f-j,\beta}^u + \hat{s}_{\delta f}^u = Q_\delta && \delta \in \mathcal{D},\ B \leq f \leq F \label{VFA_cap2}\\
    & \sum_{c\in\mathcal{C}}\sum_{j=0}^{F+B-1-f}\sum_{\beta=f-F+j+1}^B \hat{y}_{\delta c,F-j,\beta}^u + \hat{s}_{\delta f}^u = Q_\delta && \delta \in \mathcal{D},\ F+1 \leq f \leq F+B-1 \label{VFA_cap3}\\
    & \sum_{j=1}^\delta \hat{y}_{jcf\beta}^u \leq \sum_{d=1}^\delta \hat{o}_{dcf\beta}^{ukx} && \delta \in \mathcal{D}\setminus\{D\},\ c\in\mathcal{C},\ f\in\mathcal{F},\ \label{VFA_order_mix} \\[-3ex]
    & && \beta\in\mathcal{B} \notag \\
    & \sum_{\delta\in\mathcal{D}} \hat{y}_{\delta cf\beta}^u = \sum_{d\in\mathcal{D}} \hat{o}_{dcf\beta}^{ukx} && c\in\mathcal{C},\ f\in\mathcal{F},\ \beta\in\mathcal{B} \label{VFA_all} \\
    & \sum_{j=1}^{f}\sum_{\lambda=f-j+1}^{F+B-j} \hat{w}_{\delta\lambda j}^u = \hat{s}_{\delta f}^u && \delta\in\mathcal{D},\ f\in\hat{\mathcal{F}} \label{VFA_s_to_w} \\
    &\sum_{j=1}^{f-1} \hat{w}_{\delta j, f-j}^u \leq \sum_{c\in\mathcal{C}} \sum_{\beta=1}^B \hat{y}_{\delta cf\beta}^u && \delta\in\mathcal{D},\ f\in\{2,\dots,F\} \label{VFA_w_end} \\
    & \hat{w}_{\delta\lambda}^u = \sum_{f=1}^{F+B-\lambda} \hat{w}_{\delta\lambda f}^u && \delta\in\mathcal{D},\ \lambda\in\hat{\mathcal{F}} \label{VFA_wf_to_w} \\
    & \sum_{\delta\in\mathcal{D}}\sum_{f=F+1}^{F+B-1}\sum_{j=1}^{f-1} \hat{w}_{\delta j, f-j}^u = 0 && \label{VFA_to_end} \\
    & \hat{s}_{\delta f}^u,\ \hat{w}_{\delta \lambda}^u,\ \hat{w}_{\delta\lambda f}^u,\ \hat{y}_{\delta cf\beta}^u \in \mathbb{N}_0 && \delta \in \mathcal{D},\ c\in\mathcal{C},f\in\hat{\mathcal{F}},\  \label{VFA_int}\\
    & && \lambda\in\{1,\dots,F+B-f\},\ \beta\in\mathcal{B} \notag 
\end{align}

Similar to (\ref{CFA_cap1})-(\ref{CFA_all}), constraints (\ref{VFA_cap1})-(\ref{VFA_all}) ensure that we generate a feasible tentative allocation plan. Constraint (\ref{VFA_cap3}) depicts the extended time horizon of the tentative plan due to the sampled pickup times. Mirroring (\ref{CFA_s_to_w})-(\ref{CFA_wf_to_w}), constraints (\ref{VFA_s_to_w})-(\ref{VFA_wf_to_w}) guarantee the correct computation of the capacity windows. We add constraint (\ref{VFA_to_end}) to prohibit capacity windows terminating in $F,\dots,F+B-2$. In other words, capacity windows must either be followed by an occupied compartment or end with the final allocation decision in $f=F+B-1$.

Due to the longer planning horizon, we need to adapt the objectives (\ref{obj_s_first}) and (\ref{obj_w_first}):
\begin{align}
    & \max\sum_{\delta\in\mathcal{D}}\sum_{\lambda\in\hat{\mathcal{F}}} \delta\lambda \hat{w}^u_{\delta \lambda} + \sum_{\delta\in\mathcal{D}}\sum_{\lambda\in\hat{\mathcal{F}}} \frac{ (2\lambda-1)}{1+(2(F+B-1)-1)\cdot\sum_{\delta\in\mathcal{D}} Q_{\delta}} \hat{w}_{\delta \lambda}^u \label{vfa_s_first}\\
    & \max\sum_{\delta\in\mathcal{D}} \sum_{\lambda\in\hat{\mathcal{F}}} (2\lambda-1)\hat{w}^u_{\delta\lambda} + \sum_{\delta\in\mathcal{D}}\sum_{\lambda\in\hat{\mathcal{F}}}\frac{ \delta \lambda}{1+(F+B-1)\sum_{\delta\in\mathcal{D}}\delta Q_{\delta}}\hat{w}^u_{\delta \lambda} \label{vfa_w_first}
\end{align}

\subsection{Training procedure} \label{VFA_training_details}
In the subsequent paragraphs, we delve into Algorithm~\ref{alg_VFA_training}, which is an extended version of the pseudocode provided in Algorithm \ref{alg_VFA_training_short}. As a first step, we briefly (re)introduce the hyperparameters: The simulation runs for a total of $\tau^\text{max}$ days. We use $n$ to refer to the simulated decision epochs. The training procedure includes stepsizes $\alpha^\theta$ and $\alpha^{\bar{R}}$ together with $\alpha_n^{\bar{R}}$ for updating the parameter weights $\bm{\theta}$ and reward rate $\bar{R}$ respectively as well as an exploration scheme $\varepsilon_n$ for decision-making. The parameters for ER specify the size $\kappa$ of the replay memory $\mathbb{M}$ and the size $\chi$ of the samples drawn from it along with a regularization term $\gamma$ and the start $\eta$ and frequency $\zeta$ of ER updates. Lastly, we require a prespecified configuration of the CFA parameters $v_{\delta\lambda}$ as they form the foundation of the VFA features $\bm{\phi}(S_k^x)$ and govern allocation decisions during the simulation. 

\begin{algorithm}
\fontsize{9}{13}\selectfont
\caption{Training of VFA parameter weights $\bm{\theta}$}\label{alg_VFA_training}
    \textbf{Parameters:} $\alpha^{\bar{R}},\alpha^\theta, \gamma, \varepsilon_n,\zeta,\eta,\kappa,\tau^\text{max},\chi,v_{\delta\lambda}$ (for allocation decisions and computation of $\bm{\phi}(S_k^x)$)\\
    \textbf{Initialization:} $\bm{\theta}=\mathbf{0},\rho=0,\mathbb{M}=\{\},n=0,\bar{R}=0, S_0^x$
\begin{algorithmic}[1]
\For{$\tau=1,\dots,\tau^{\text{max}}$}
    \Repeat
        \State $n \gets n+1$
        \State sample exogenous information $W_n$ to obtain $S_n = S^{MW}(S^x_{n-1},W_n)$
        \If{$t_n\leq T$}
            \State make $\varepsilon$-greedy demand control decision $g_n$ to obtain $S_n^x=S^{Mx}(S_n,g_n)$
            \State compute TD error $\Delta=R(S_n,g_n)-\bar{R}+\hat{V}(S_n^x|\bm{\theta}_{n-1})-\hat{V}(S_{n-1}^x|\bm{\theta}_{n-1})$
            \State update reward rate $\rho\gets\rho+\alpha^{\bar{R}}(1-\rho),\ \alpha^{\bar{R}}_n\gets\frac{\alpha^{\bar{R}}}{\rho},\ \bar{R}\gets\bar{R}+\alpha^{\bar{R}}_n\Delta$
            \State perform TD update of parameter weights $\bm{\theta}\gets\bm{\theta}+\alpha^\theta\Delta\bm{\phi}(S^x_{n-1})$
            \State update memory $\mathbb{M}\gets\mathbb{M}\cup\{n\}$ and store experience $(\bm{\phi}_n^0,\bm{\phi}_n^\text{reject},\bm{\phi}_n^\text{accept},R_n^\text{accept})$
            \If{$|\mathbb{M}|>\kappa$}
                \State remove oldest experience from memory $\mathbb{M}$
            \EndIf
        \EndIf
    \Until{$t_n=T+1$}
    \State make allocation decision $A_n$ to obtain $S_n^x=S^{Mx}(S_n,A_n)$
    \If{$(\tau_n>\eta)\wedge(\tau_n\bmod\zeta=0)$}
        \State uniformly draw sample $\bar{\mathbb{M}}$ from $\mathbb{M}$ with $|\bar{\mathbb{M}}|=\min\{\chi,|\mathbb{M}|\}$ \label{ER_start}
        \State compute ER update targets $\nu_i\gets\max\{R_i^\text{accept}-\bar{R}+\bm{\theta}^T\bm{\phi}_i^\text{accept};-\bar{R}+\bm{\theta}^T\bm{\phi}_i^\text{reject}\}\ \forall i\in\bar{\mathbb{M}}$ \label{alg_ER_update_targets}
        \State perform ER update on $\bm{\theta}$ via ridge regression using $\bar{\mathbb{M}}$ and $\gamma$ \label{ER_end}
    \EndIf
\EndFor
\end{algorithmic}
\end{algorithm}

We initialize the algorithm by setting all parameter weights $\bm{\theta}$ and our estimate for the reward rate $\bar{R}$ to zero. This corresponds to the myopic policy where decisions are made exclusively based on the immediate reward without anticipating potential downstream effects of decisions. Furthermore, the algorithm starts with an empty replay memory $\mathbb{M}=\{\}$ and in decision epoch $n=0$ with an initial post-decision state $S_0^x$. The latter can be constructed, e.g., by starting with an empty locker and no pending orders and simulating a few days with the myopic policy. Finally, $\rho$ is part of the estimation for $\bar{R}$ and initialized with zero.

During each simulated day, we generate the next decision epoch by sampling exogenous information $W_n$ (Section \ref{sdp_transitions}) and transitioning into a new pre-decision state $S_n$. In case of a request arrival, we make an $\varepsilon$-greedy demand control decision, i.e., we choose the decision that is deemed optimal according to our current parameter weights $g^*_n=\arg\,\max_{g\in\mathcal{G}(S_n)}\{R(S_n,g)+\hat{V}(S^{Mx}(S_n,g)|\bm{\theta}_{n-1})\}$ with probability $1-\varepsilon_n$ and select a random decision out of $\mathcal{G}(S_n)$ with probability $\varepsilon_n$.

After demand control, the system evolves into post-decision state $S_n^x$. Based on this transition, we can compute the TD error $\Delta$ to update our estimate $\bar{R}$ as well as the parameter weights $\bm{\theta}$ \citep[Chapters 2.6 and 10]{sutton+barto}. The data collected during decision epoch $n$ is stored as an experience $(\bm{\phi}_n^0,\bm{\phi}_n^\text{reject},\bm{\phi}_n^\text{accept},R_n^\text{accept})$ that encompasses the feature vector of the previous post-decision state $\bm{\phi}_n^0=\bm{\phi}(S_{n-1}^x)$ as well as the feature vector resulting from rejection of the current request $\bm{\phi}_n^\text{reject}=\bm{\phi}(S^{Mx}(S_n,0))$. Moreover, if the request in $n$ is feasible, we include the post-decision state features resulting from acceptance $\bm{\phi}_n^\text{accept}=\bm{\phi}(S^{Mx}(S_n,1))$ and the reward $R_n^\text{accept}=m_{c_{r_n}}$. For an infeasible request, we set  $\bm{\phi}_n^\text{accept}=\mathbf{0}$ and $R_n^\text{accept}=-\infty$. We track the indices of decision epochs currently saved as experience in the replay memory $\mathbb{M}$. If the size of the replay memory exceeds $\kappa$, we delete the oldest experience.

The sequence of demand control decisions and TD updates of the VFA parameter weights $\bm{\theta}$ carries on until the day concludes with the allocation decision in $T+1$, which is shaped by the CFA parameterization $v_{\delta\lambda}$. In addition, after the first $\eta$ days, we perform an ER-based update for $\bm{\theta}$ every $\zeta$ days. The threshold $\eta$ serves to ensure that a sufficient amount of experience has accumulated in the replay memory.

For the ER update, we generate a simple random sample without replacement from $\mathbb{M}$, which we denote by $\bar{\mathbb{M}}$. The sample contains $\chi$ previously encountered decision epochs (if $\chi>|\mathbb{M}|$, $\bar{\mathbb{M}}=\mathbb{M}$), each associated with a stored experience $(\bm{\phi}_n^0,\bm{\phi}_n^\text{reject},\bm{\phi}_n^\text{accept},R_n^\text{accept})$. Based on this data, we can compute prediction targets $\nu_i\ \forall i\in\bar{\mathbb{M}}$ as shown in Line \ref{alg_ER_update_targets} in Algorithm \ref{alg_VFA_training} and use them in a ridge regression to obtain updated parameter weights $\bm{\theta}$. Using standardized features $\bar{\bm{\phi}}_i^0$ and ridge parameter $\gamma$, we obtain $\bm{\theta}^{ER^*}=\arg\,\min_{\bm{\theta}^{ER}}\sum_{i\in\bar{\mathbb{M}}}[\nu_i-\bm{\theta}^{ER^T}\bar{\bm{\phi}}_i^0]+\gamma\sum_{\delta\in\mathcal{D}}\sum_{\lambda\in\hat{\mathcal{F}}}(\theta_{\delta\lambda}^{ER})^2$ and transform $\bm{\theta}^{ER^*}$ into $\bm{\theta}$ by reversing the standardization. 

As we approximate post-decision state values, the ER update is less off-policy than its standard implementation in RL. More specifically, the update targets are more on-policy as we generate them by making the decision in the replayed decision epoch that is considered optimal according to our current parameter weights. The distribution of experiences in the replay memory depends on previous parameter weights and is hence off-policy.

\subsection{Structure-enforcement during experience replay} \label{VFA_enf_details}
To guide the learning process and enforce structure in the value function, we impose the following constraints on the experience replay update:
\begin{align}
    &\theta_{\delta\lambda}\geq\theta_{i\lambda} && \delta\in\mathcal{D},\ i\in\{1,\dots,\delta-1\},\ \lambda\in\hat{\mathcal{F}} \label{structure_size} \\
    & \theta_{\delta\lambda}\geq\theta_{\delta j} && \delta\in\mathcal{D},\ \lambda\in\hat{\mathcal{F}},\ j\in\{1,\dots,\lambda-1\} \label{structure_length} \\
    & \theta_{\delta\lambda}\geq 0 && \delta\in\mathcal{D},\ \lambda\in\hat{\mathcal{F}} \label{theta_nonneg}
\end{align}

Constraints (\ref{structure_size}) require the valuation of a capacity window of a given length $\lambda$ to be nondecreasing with increasing compartment size $\delta$. In complementary fashion, constraints (\ref{structure_length}) ensure that the valuation of a capacity window with given compartment size $\delta$ is nondecreasing with increasing window length $\lambda$. Lastly, constraints (\ref{theta_nonneg}) guarantee nonnegativity of the parameter weights.

\section{Truncated online rollout} \label{a_rollout}
After completing the learning process, the VFA can be directly applied on its own for demand control by substituting the value function in (\ref{optimal_dc}) with the value estimates $\hat{V}(S|\bm{\theta})$. As an optional enhancement, we can further improve demand control by combining the offline trained value function with a truncated online rollout \citep{bertsekas2022}. More specifically, starting from a post-decision state, we generate $\Omega$ sample paths by simulating ahead into the future for a limited horizon $\xi$ using a base policy. For the base policy, one can use the VFA or, to save computational effort, a simple myopic policy that accepts every feasible request. The terminal state of each sample path is then evaluated using the VFA. Computing the sum of rewards collected during the simulation as well as the value of the terminal state and taking the average across the sample paths, we obtain a refined estimate for the post-decision state values.

In general, VFA is always associated with aggregation and hence a loss of information contained in the state variable. In our case, this is especially relevant given that the features of our VFA are based on tentative allocation plans that consider days on an aggregate level, thereby neglecting individual points in time $t$. As a consequence, our features contain no explicit information on the amount of time remaining until the next allocation decision. It is merely included implicitly in the sampled values $\beta$ of the number of days after allocation until the pickup occurs for orders currently in the locker as the pickup probabilities are conditioned on the dwell time and current time $t$. The truncated rollout compensates this through its very detailed consideration of the dynamics and the near future.

\section{Policies: Implementation details} \label{policy_details}
In the following, we provide details for some of the policies used in our experiments.

\subsection{Bottom-up allocation}
For the allocation according to BU, we combine the CFA decision space formulated in Appendix \ref{cfa_decision_space} with the following objective function: 
\begin{equation} \label{BU_obj}
    \max \sum_{\delta\in\mathcal{D}} \delta s_{\delta ,f=1} \notag
\end{equation}
BU exclusively considers the next allocation decision ($f=1$) and maximizes the available capacity in hierarchical order of the compartment size with the largest compartment size getting the highest priority. In other words, the objective prioritizes maximizing the number of empty compartments with the maximum size $D$ in $f=1$ as a primary objective, followed by maximizing the number of empty compartments with size $D-1$ in $f=1$ as a secondary objective, and so forth. 

\subsection{Deterministic linear program}
The DLP is a fundamental concept in the revenue management literature and, for a problem similar to the DPLDMP, also applied in practice by retailer Amazon \citep{sethuraman_etal}. Accordingly, we test a DLP-based approach for demand control in our computational study to see how our proposed solution method fares against it. Since we consider a novel problem variant involving uncertain usage durations and upgrading, we cannot directly apply existing versions of the DLP and must first transfer the concept to our setting by formulating an adequate linear program. 

Our DLP approach is based on certainty equivalent control \citep{bertsimas_popescu_2003} and works as follows: To make a demand control decision in pre-decision state $S_k=(\tau_k,t_k,L_k,O_k,r_k)$, we solve two instances of the linear program formalized by (\ref{dlp_obj})-(\ref{dlp_nonnegative}). One of these instances represents acceptance of the current request $r_k$, the other rejection. The difference in the optimal objective values then serves as an estimate for the opportunity cost, which allows us to apply the decision rule in (\ref{optimal_dc}).

The DLP covers a horizon of $\varphi^\text{DLP}$ days, which is the only hyperparameter of this approach. Note that days are modeled on an aggregate level, i.e., we neglect individual points in time. We use $\vartheta,\varphi\in\{1,\dots,\varphi^\text{DLP}\}$ to denote days in the DLP with $\vartheta,\varphi=1$ corresponding to the current day $\tau_k$. The existing pending orders are represented by $\Tilde{O}_k$. We model the acceptance of the current request $r_k$ by modifying $O_k$ to $\Tilde{O}_k$ with $\Tilde{o}^k_{d c f}=o^k_{d c f}+\mathbf{1}(d=d_{r_k}, c=c_{r_k},f=f_{r_k})$ and $\mathbf{1}(\cdot)$ symbolizing the indicator function. For the second instance, which results from rejecting $r_k$, we set $\Tilde{O}_k=O_k$.
\newpage
The probabilistic information on request arrivals and pickups is incorporated through the following three parameters:
\begin{itemize}
    \item $\Hat{o}_{dc\varphi}$ equals the expected number of future requests with parcel size $d$, customer type $c$, and delivery date $\varphi$.
    \item $\Bar{p}_{ch\varphi}$ denotes the conditional probability that a parcel belonging to a customer of type $c$ with dwell time $h$ on day $\tau_k$ is still in the locker for at least one point in time on day $\varphi$. In this way, we include the pickups for orders currently occupying the locker $L_k$.
    \item To incorporate pickups of pending orders $\Tilde{O}_k$ and future accepted requests, we let $\Hat{p}_{c\vartheta\varphi}$ represent the probability that a parcel of customer type $c$ allocated on day $\vartheta$ is still in the locker for at least one point in time on day $\varphi$.
\end{itemize}

Lastly, we define decision variable $\Tilde{y}_{\delta c\varphi}$ as the number of compartments of size $\delta$ to which we allocate a parcel belonging to a customer of type $c$ on day $\varphi$.

We formulate the DLP as follows:
\begin{align}
    & \max \sum_{c\in\mathcal{C}}\sum_{\delta\in\mathcal{D}}m_c\left(\sum_{\varphi=1}^{\varphi^\text{DLP}}\Tilde{y}_{\delta c\varphi} - \sum_{f\in\mathcal{F}}\Tilde{o}_{\delta cf}\right) && \label{dlp_obj}\\
    & \sum_{c\in\mathcal{C}} \sum_{j=1}^\varphi \Hat{p}_{cj \varphi}\Tilde{y}_{\delta c j} + \sum_{c\in\mathcal{C}} \sum_{h\in\mathcal{B}} \bar{p}_{ch\varphi}l_{\delta c h}^k \leq Q_\delta && \delta \in \mathcal{D},\ \varphi\in\{1,\dots,\varphi^\text{DLP}\} \label{dlp_cap}\\
    & \sum_{d=1}^\delta \Tilde{y}_{jc\varphi} \leq \sum_{j=1}^\delta (\Tilde{o}_{dc\varphi}^k+\Hat{o}_{dc\varphi}) && \delta \in \mathcal{D},\ c\in\mathcal{C},\ \varphi\in\mathcal{F} \label{dlp_order_mix1} \\
    & \sum_{j=1}^\delta \Tilde{y}_{jc\varphi} \leq \sum_{j=1}^\delta \Hat{o}_{jc\varphi} && \delta \in \mathcal{D},\ c\in\mathcal{C},\ F<\varphi\leq\varphi^\text{DLP} \label{dlp_order_mix2} \\
    & \sum_{\delta\in\mathcal{D}} \Tilde{y}_{\delta cf} \geq \sum_{d\in\mathcal{D}} \Tilde{o}_{dcf}^k && c\in\mathcal{C},\ f\in\mathcal{F} \label{dlp_all} \\
    & \Tilde{y}_{\delta c\varphi} \geq 0 && \delta \in \mathcal{D},\ c\in\mathcal{C},\ \varphi\in\{1,\dots,\varphi^\text{DLP}\} \label{dlp_nonnegative}
\end{align}

The objective (\ref{dlp_obj}) maximizes the number of future accepted (and thus allocated) requests weighted by their priority. Note that $\Tilde{y}_{\delta c\varphi}$ also encompasses the allocations for existing pending orders in $\Tilde{O}_k$, which is why we must subtract these orders in the objective. Constraints (\ref{dlp_cap}) ensure that the number of compartments is not exceeded, taking into account the current occupancy of the locker, the allocation of pending and future orders over time as well as the pickups. Constraints (\ref{dlp_order_mix1}) and (\ref{dlp_order_mix2}) guarantee that the orders are assigned to compatible compartments and that the number of allocated parcels does not exceed the number of pending orders and the expected demand-to-come. Constraints (\ref{dlp_all}) state that all orders in $\Tilde{O}_k$ must be allocated.

\section{Hyperparameters and instance generation} \label{cs_inst_generation}
In the following, we state the hyperparameter values used during our experiments. For the VFA training, we employ the following calibration: To generate the initial state $S_0^x$, we first start with an empty locker and no pending orders and simulate 5 days with the myopic demand control approach FC. The training procedure then runs for $\tau^\text{max}=2500$ days. For the update stepsizes, we set $\alpha^\theta=\alpha^{\bar{R}}=0.001$. Regarding exploration, we start with $\varepsilon_1=1$ and linearly decay it to 0.1 until day $\tau=1200$, after which it is kept constant for the remainder of the training procedure. The ER parameters for (R)V-CER and (R)V-ER include $\kappa=10,000$ for the size of the replay memory and $\chi=4,000$ for the sample size along with regularization term $\gamma=4$, start $\eta=100$, and frequency $\zeta=5$. To calculate the features, we use $U=10$ scenarios. The truncated online rollout uses $\Omega=5$ sample paths by simulating ahead into the future for a horizon of $\xi=5$ points in time with the myopic demand control approach FC as the base policy. For the DLP, we tested a horizon $\varphi^\text{DLP}$ of 5, 10, 15, and 20 days and report the results for the best-performing one, i.e., $\varphi^\text{DLP}=5$.

In pretests, we additionally considered another variation of VFA training where, in the spirit of the `Combined-Q' algorithm by \cite{zhang+sutton2017}, both the TD and ER update are performed in every decision epoch. However, this approach led to numerical instabilities and divergence. It is hence not included in the computational study.

To test our policies, we generated 30 simulated streams of incoming customer requests spanning 40 days each. For the sake of comparability, we used these 30 streams of customer arrivals for each of our nine settings such that the settings exclusively differ in the priority weight of premium customers and the pickup distributions and not in the sequence of incoming requests. For every setting, we ran each of our 27 policies on each of the 30 customer streams starting with an initially empty locker and no pending orders and discarded the first ten days to eliminate temporary effects caused by the initial state of the system. 

Given that the VFA training is based on simulation and hence subject to uncertainty, we trained every VFA five times and report the average results. The coefficient of variation in the objective value (on average for all instances and settings) was below 3\% for each of our VFA variants, meaning that the learning process produces stable results.

\section{Mismatching feature calculation and allocation} \label{mismatch}
With this supplementary experiment, we investigate how a mismatch between the objective for the VFA feature calculation and the allocation decision affects performance. More precisely, we combine two different objective functions, one for computing the tentative allocation plans that form the basis of the VFA features and another for the allocation decision. Using this configuration, we train the VFA for RV-CER accordingly. Note that we do not alter the feature or allocation objective after the training, i.e., the VFA is trained on the mismatch. Given that we can choose from three objectives (DL, LD, and BU) and have already investigated the cases where the objectives for the features and allocation are identical, we obtain a total of six additional combinations and hence VFAs to train. To keep the computational effort moderate, we slightly deviate from the setup used for the main study and evaluate the policies on 15 instances (instead of 30) spanning 25 days each (instead of 40) and train each VFA three times (instead of five), but everything else is kept constant. We report the objective improvement compared to our benchmark $\pi^\text{FC}_\text{DL}$ in Table \ref{res_mismatch}.

\begin{table}%{r}{4.5cm}
\centering
\renewcommand{\arraystretch}{1.2}
\caption{Objective improvement of $\pi^\text{RV-CER}$ over $\pi^\text{FC}_\text{DL}$ in \%.} \label{res_mismatch}
\resizebox{0.3\textwidth}{!}{\begin{tabular}{>{\centering\arraybackslash}p{1.5cm} >{\centering\arraybackslash}p{0.7cm} >{\centering\arraybackslash}p{1.2cm}>{\centering\arraybackslash}p{1.2cm}>{\centering\arraybackslash}p{1.2cm}} \hline
 & & \multicolumn{3}{c}{Allocation} \\ \cline{3-5}
         &    & DL    & LD   & BU \\ \hline
         & DL & 13.7  & 13.8  & -42.8 \\
Features & LD & 13.7  & 13.7 & -42.8 \\
         & BU & -47.5 & -47.4  & 5.9 \\
\hline
\end{tabular}}
\end{table}

We observe that if we combine DL and LD, the performance remains basically identical (regardless of which objective is employed for feature calculation or allocation). By contrast, if we apply BU in combination with DL or LD, the resulting policies perform drastically worse than the benchmark policy. Moreover, using BU for the features and DL or LD for the allocation produces slightly worse results than vice versa.

To explain these results, we first point out that DL and LD share strong similarities as both allocation schemes involve a lexicographic optimization that is based on the same two objectives and merely differs in their hierarchical ordering. Additionally, Figure \ref{figure_performance} shows that DL and LD perform very similar for all tested demand control approaches. Meanwhile, VFA generally has a tendency to perform worse with BU compared to DL and LD even if the objectives for feature calculation and allocation match. At its core, BU is fundamentally different to DL and LD since it does not draw on the concept of capacity windows and instead focuses entirely on the next allocation decision.

From these findings, we infer that the training procedure of the VFA is able to handle and compensate a mismatch between feature calculation and allocation objectives if the difference is small. However, VFA cannot overcome the impact of two drastically different objectives even if trained on them. If the allocation decisions are made in a radically different fashion than the anticipated decisions in the tentative allocation plans for the features, the features become an unreliable estimate of the available capacity, which makes learning a meaningful evaluation in the form of adequate feature weights difficult to nearly impossible. In conclusion, it is crucial to ensure that the two solution components for demand control and allocation align well.

\section{Locker occupancy and pending orders} \label{occupancy+orders}
\begin{figure} 
\centerline{\includegraphics[scale = 0.55]{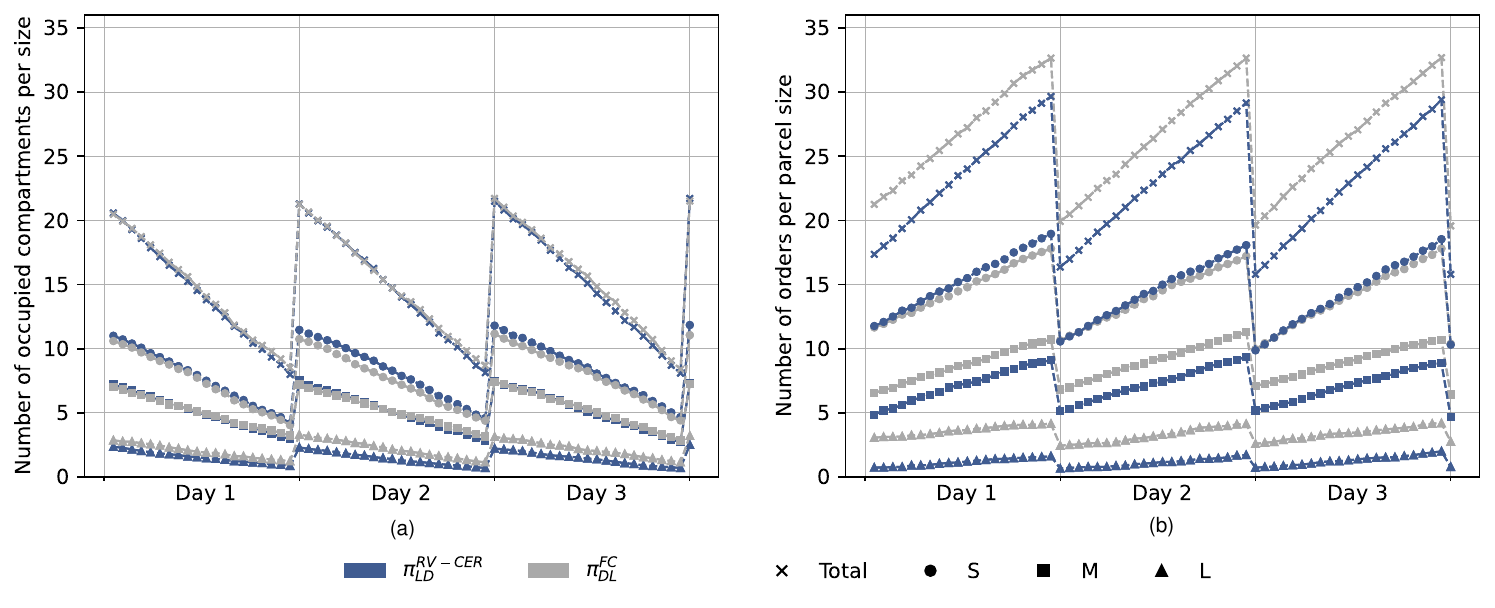}} 
\caption{Number of occupied locker compartments and pending orders. \label{figure_nr_orders}} 
{} 
\end{figure}

To gain an understanding of the system's state and how it evolves over time, we track the number of occupied locker compartments and pending orders over three randomly selected consecutive days. As expected, the number of occupied compartments is highest at the end of each day immediately after the allocation decision and then decreases over the course of each day as customers collect their parcels. The pending orders behave in a complementary fashion: With the allocation decision, pending orders scheduled for delivery turn into orders occupying the locker, leading to a drop in the number of pending orders directly after allocation. During each day, the number of pending orders rises again due to accepted requests. As shown in Figure \ref{figure_nr_orders}, the number of occupied compartments behaves similarly for the two policies, but $\pi^\text{FC}_\text{DL}$ has more pending orders in the system at every point in time since it tends to accept more requests with longer lead times. In line with our findings in Section \ref{parcel_size}, we observe that for small parcels, $\pi^\text{RV-CER}_\text{LD}$ has a similar number of pending orders to $\pi^\text{FC}_\text{DL}$, but tends to have less pending orders with medium and large parcels.

\end{appendices}

\bibliographystyle{plainnat}
\bibliography{main} 
\end{document}